\documentclass[10pt,twocolumn,letterpaper]{article}

\usepackage{cvpr}
\usepackage{times}
\usepackage{epsfig}
\usepackage{graphicx}
\usepackage{amsmath}
\usepackage{amssymb}
\usepackage{xcolor}
\usepackage{booktabs}
\usepackage{afterpage}
\usepackage[]{subfig}
\usepackage{balance}

\usepackage{authblk}
\makeatletter
\renewcommand\AB@affilsepx{, \protect\Affilfont}
\makeatother

\newcommand{\smplHF}{\mbox{SMPL-X}\xspace}
\newcommand{\smplifyPP}{\mbox{SMPLify-X}\xspace}

\newcommand{\smpl}{\mbox{SMPL}\xspace}
\newcommand{\smplH}{\mbox{SMPL+H}\xspace}
\newcommand{\mano}{\mbox{MANO}\xspace}
\newcommand{\flame}{\mbox{FLAME}\xspace}
\newcommand{\smplify}{\mbox{SMPLify}\xspace}
\newcommand{\frank}{\mbox{Frank}\xspace}
\newcommand{\vposer}{\mbox{VPoser}\xspace}
\newcommand{\caesar}{\mbox{CAESAR}\xspace}

\newcommand{\facewarehouse}{\mbox{FaceWarehouse}\xspace}
\newcommand{\openDR}{\mbox{OpenDR}\xspace}
\newcommand{\chumpy}{\mbox{Chumpy}\xspace}
\newcommand{\openpose}{\mbox{OpenPose}\xspace}
\newcommand{\cuda}{\mbox{CUDA}\xspace}
\newcommand{\pytorch}{\mbox{PyTorch}\xspace}
\newcommand{\mocap}{\mbox{MoCap}\xspace}

\newcommand{\inthewild}{in-the-wild\xspace}
\newcommand{\twoD}{2D\xspace}
\newcommand{\threeD}{3D\xspace}
\newcommand{\gt}{ground-truth\xspace}
\newcommand{\facs}{\mbox{FACS}\xspace}

\newcommand{\mExpr}{\mathcal{E}}
\newcommand{\mColl}{\mathcal{C}}
\newcommand{\Jest}{J_{est}}
\newcommand{\vertex}{v}
\newcommand{\normal}{n}
\newcommand{\para}{\theta}

\newcommand{\supmat}{Sup.~Mat.\xspace}

\newcommand{\citeMOSH}{\cite{lopermahmoodetal2014,amass2019}\xspace}

\newcommand{\websiteSMPLX}{\mbox{\url{https://smpl-x.is.tue.mpg.de}}}

\newcommand\blfootnote[1]{%
  \begingroup
  \renewcommand\thefootnote{}\footnote{#1}%
  \addtocounter{footnote}{-1}%
  \endgroup
}

\usepackage[pagebackref=true,breaklinks=true,letterpaper=true,colorlinks,bookmarks=false]{hyperref}
\usepackage[final]{pdfpages}

\def\cvprPaperID{****}

\cvprfinalcopy

\ifcvprfinal\fi
\begin{document}

\title{Expressive Body Capture: \threeD Hands, Face, and Body from a Single Image\vspace*{-1.5em}}

\author[1,2]{Georgios Pavlakos\textsuperscript{*}}
\author[1]{Vasileios Choutas\textsuperscript{*}}
\author[1]{Nima Ghorbani}
\author[1]{Timo Bolkart}
\author[1]{Ahmed A. A. Osman}
\author[1]{Dimitrios Tzionas}
\author[1]{Michael J. Black}
\affil[1]{\normalsize						MPI for Intelligent Systems, T{\"u}bingen, DE 	 }
\affil[2]{									University of Pennsylvania, PA, USA
											\authorcr 	{\tt \small \{gpavlakos, vchoutas, nghorbani, tbolkart, aosman, dtzionas, black\}@tuebingen.mpg.de}
}

\maketitle
\thispagestyle{empty}
\vspace*{-2.5em}

\begin{abstract}
\vspace*{-0.6em}
To facilitate the analysis of human actions, interactions and emotions, we compute a \threeD model of human body pose, hand pose, and facial expression from a single monocular image.
To achieve this, we use thousands of \threeD scans to train a new, unified, \threeD model of the human body, {\em \smplHF}, that extends \smpl with fully articulated hands and an expressive face.
Learning to regress the parameters of \smplHF directly from images is challenging without paired images and \threeD ground truth.
Consequently, we follow the approach of \smplify, which estimates \twoD features and then optimizes model parameters to fit the features.
We improve on \smplify in several significant ways:
(1) we detect \twoD features corresponding to the face, hands, and feet and fit the full \smplHF model to these;
(2) we train a new neural network pose prior using a large \mocap dataset;
(3) we define a new interpenetration penalty that is both fast and accurate;
(4) we automatically detect gender and the appropriate body models (male, female, or neutral);
(5) our \pytorch implementation achieves a speedup of more than $8\times$ over \chumpy.
We use the new method, {\em \smplifyPP}, to fit \smplHF to both controlled images and images in the wild. 
We evaluate \threeD accuracy on a new curated dataset comprising $100$ images with pseudo \gt.
This is a step towards automatic expressive human capture from monocular RGB data.
The models, code, and data are available for research purposes at \websiteSMPLX.
\vspace*{-0.6em}
\blfootnote{$^*$ equal contribution}
\end{abstract}

\vspace*{-0.5em}
\section{Introduction}
Humans are often a central element in images and videos. 
Understanding their posture, the social cues they communicate, and their interactions with the world is critical for holistic scene understanding.
Recent methods have shown rapid progress on estimating the major body joints, hand joints and facial features in \twoD \cite{cao2017realtime,insafutdinov2016deepercut,simon2017hand}. 
Our interactions with the world, however, are fundamentally \threeD and recent work has also made progress on the \threeD estimation of the major joints and rough \threeD pose directly from single images \cite{bogo2016keep,kanazawa2018end,omran2018neural,pavlakos2018learning}. 

\newcommand{\ImgEmotionsHeightXX}{0.246}
\newcommand{\ImgEmotionsHeightYY}{0.31}
\newcommand{\ImgEmotionsHeightZZ}{0.160}
\newcommand{\ImgEmotionsHSpaceXX}{-3.0mm}
\newcommand{\ImgEmotionsVSpaceXX}{-4.0mm}
\begin{figure}
	\centering
	\vspace*{-01.00em}
	\subfloat{	\includegraphics[trim=00mm 00mm 00mm 00mm,   clip=true, height=\ImgEmotionsHeightYY \textwidth]{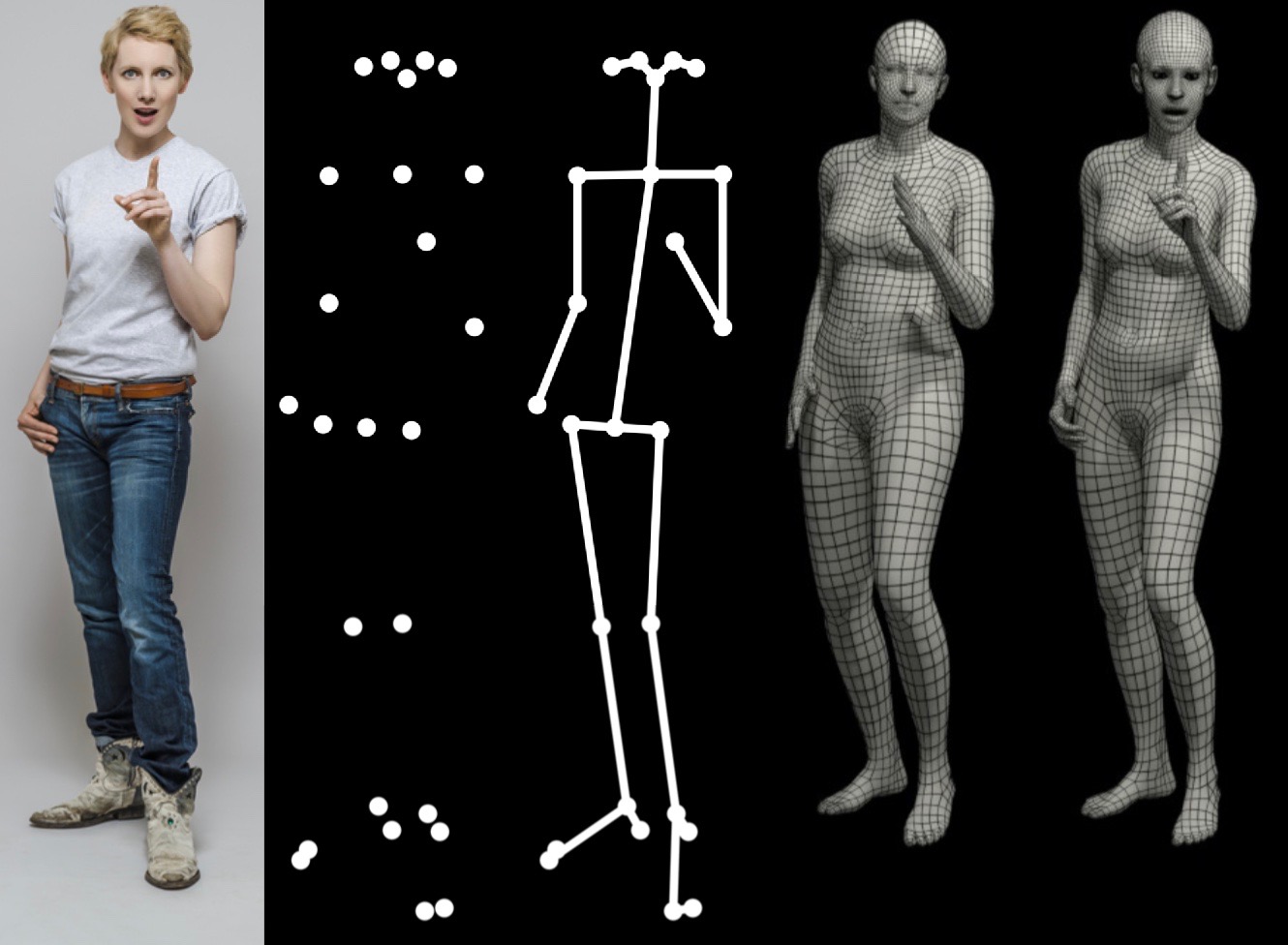}}
	\vspace*{-00.50em}
	\caption{
				Communication and gesture rely on the \emph{body} pose, \emph{hand} pose, and \emph{facial} expression, all \emph{together}. 
				The major joints of the body are not sufficient to represent this and current \threeD models are not expressive enough. 
				In contrast to prior work, our approach estimates a more detailed and expressive \threeD model from a single image. 
				From left to right: RGB image, major joints, skeleton, \smpl (female), \smplHF (female). 
				The hands and face in \smplHF enable more \emph{holistic} and \emph{expressive} body capture. 
	}
	\label{fig:fromDotsToModels}
\vspace*{-3mm}
\end{figure}

To understand human behavior, however, we have to capture more than the major joints of the body -- we need the full \threeD surface of the body, hands and the face.
There is no system that can do this today due to several major challenges including the lack of appropriate \threeD models and rich \threeD training data.
Figure \ref{fig:fromDotsToModels} illustrates the problem. 
The interpretation of expressive and communicative images is difficult using only sparse \twoD information or \threeD representations that lack hand and face detail. 
To address this problem, we need two things. 
First, we need a \threeD model of the body that is able to represent the complexity of human faces, hands, and body pose.
Second, we need a method to extract such a model from a single image.

\newcommand{\ImgEmotionsHeightCA}{0.240}
\newcommand{\ImgEmotionsHeightCB}{0.246}
\newcommand{\ImgEmotionsHeightCC}{0.249}
\newcommand{\ImgEmotionsHSpaceCC}{-3.0mm}
\newcommand{\ImgEmotionsVSpaceCC}{-4.0mm}
\begin{figure*}
	\centering
	\subfloat{	\includegraphics[trim=00mm 00mm 00mm 00mm,   clip=true, width=0.96 \textwidth]{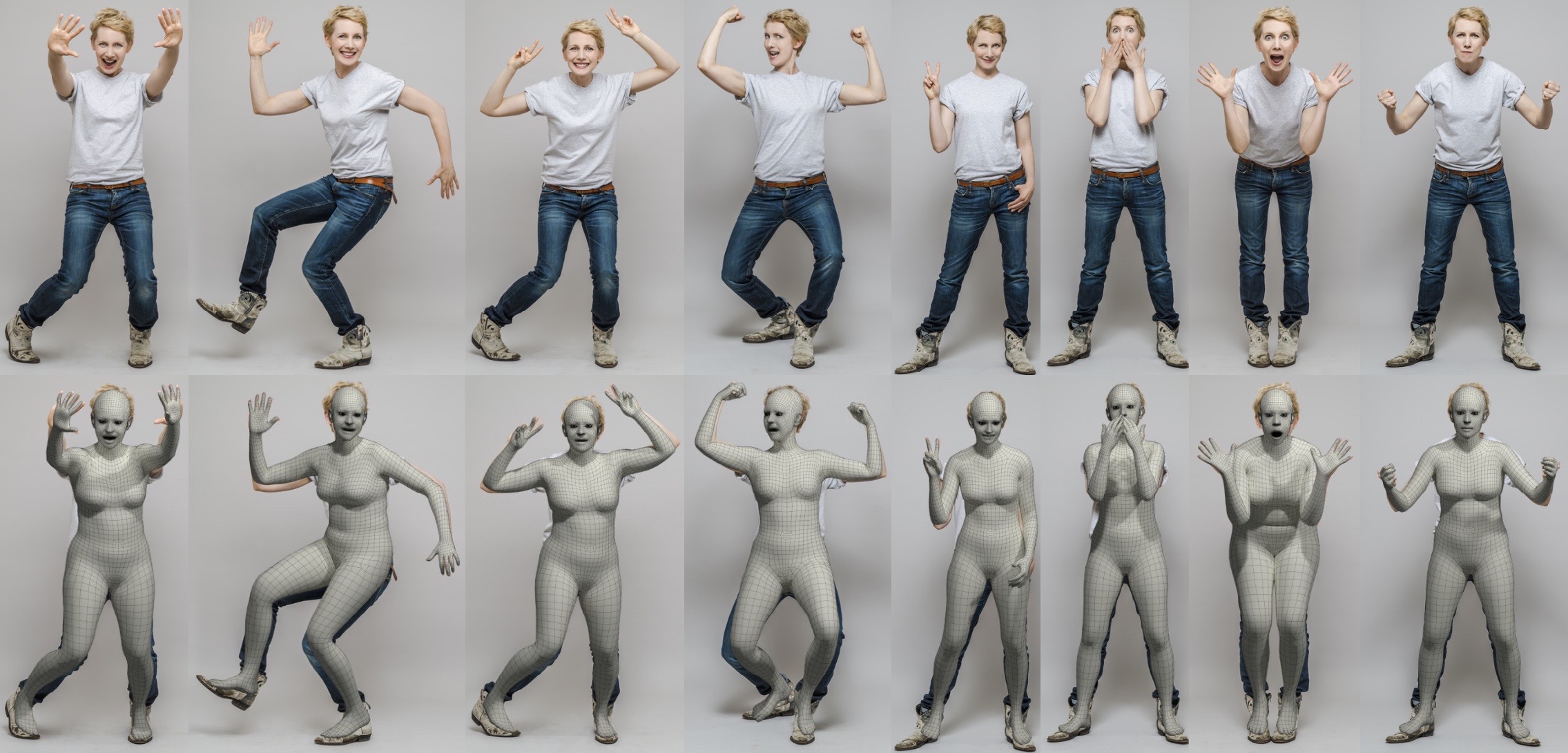}		}
	\vspace*{-02.00mm}
	\caption{
				We learn a new \threeD model of the human body called \emph{\smplHF} that jointly models the human body, face and hands. 
				We fit the female \emph{\smplHF} model with \emph{\smplifyPP} to single RGB images and show that it captures a rich variety of \emph{natural} and \emph{expressive} \threeD human poses, gestures and facial expressions.  
	}
	\label{fig:emotionsGETTY}
\vspace*{-3mm}	
\end{figure*}

Advances in neural networks and large datasets of manually labeled images have resulted in rapid progress in \twoD human ``pose'' estimation.
By ``pose'', the field often means the major joints of the body.  
This is not sufficient to understand human behavior as illustrated in Fig.~\ref{fig:fromDotsToModels}.
OpenPose \cite{cao2017realtime,OpenPoseWEB,simon2017hand} expands this to include the \twoD hand joints and \twoD facial features.
While this captures much more about the communicative intent, it does not support reasoning about surfaces and human interactions with the \threeD world.

Models of the \threeD body have focused on capturing the overall shape and pose of the body, excluding the hands and face \cite{allen2003space,Allen:2006:LCM,Anguelov05,hasler2009statistical,SMPL:2015}. 
There is also an extensive literature on modelling hands \cite{MSR_2015_CVPR_learnShapeModel,Melax:2013:handPhysics,Lepetit:ICCV:2015:handCnnLoop,OikonomidisBMVC,romero2017embodied,DART-Schmidt-RSS-14,srinath_iccv2013,Tkach:SIGGRAPH:2016,Tzionas:IJCV:2016} 
and faces \cite{Amberg2008,BlanzVetter1999,Booth2017,Brunton2014_Review,Cao2014_FaceWarehouse,li2017learning,BFM2009,Vlasic2005,Yang2011} in \threeD but in isolation from the rest of the body.
Only recently has the field begun modeling the body together with hands \cite{romero2017embodied}, or together with the hands and face \cite{joo2018total}.
The \frank model \cite{joo2018total}, for example, combines a simplified version of the \smpl body model \cite{SMPL:2015}, with an artist-designed hand rig, and the \facewarehouse \cite{Cao2014_FaceWarehouse} face model.
These disparate models are stitched together, resulting in a model that is not fully realistic.

Here we learn a new, holistic, body model with face and hands from a large corpus of \threeD scans.
The new \emph{\smplHF} model (\emph{\smpl eXpressive}) is based on \smpl and retains the benefits of that model: compatibility with graphics software, simple parametrization, small size, efficient, differentiable, etc.
We combine \smpl with the \flame head model \cite{li2017learning} and the \mano hand model \cite{romero2017embodied} and then register this combined model to $5586$ \threeD scans that we curate for quality.
By learning the model from data, we capture the natural correlations between the shape of bodies, faces and hands and the resulting model is free of the artifacts seen with \frank.
The expressivity of the model can be seen in Fig.~\ref{fig:emotionsGETTY} where we fit \smplHF to  expressive RGB images, as well as in Fig.~\ref{fig:inthewild_success} where we fit \smplHF to images of the public LSP dataset \cite{johnson2010clustered}. 
\smplHF is freely available for research purposes.

Several methods use deep learning to regress the parameters of \smpl from a single image \cite{kanazawa2018end,omran2018neural,pavlakos2018learning}. 
To estimate a \threeD body with the hands and face though, there exists no suitable training dataset. 
To address this, we follow the approach of \smplify. 
First, we  estimate \twoD image features ``bottom up'' using \openpose \cite{cao2017realtime,simon2017hand,wei2016convolutional}, which detects the joints of the body, hands, feet, and face features.
We then fit the \smplHF model to these \twoD features ``top down'', with our method called \emph{\smplifyPP}. 
To do so, we make several significant improvements over \smplify.
Specifically, we learn a new, and better performing, pose prior from a large dataset of motion capture data \citeMOSH using a variational auto-encoder.
This prior is critical because the mapping from \twoD features to \threeD pose is ambiguous.
We also define a new (self-) interpenetration penalty term that is significantly more accurate and efficient than the approximate method in \smplify; it remains differentiable.
We train a gender detector and use this to automatically determine what body model to use, either male, female or gender neutral.
Finally, one motivation for training direct regression methods to estimate \smpl parameters is that \smplify is slow.
Here we address this with a \pytorch implementation that is at least $8$ times faster than the corresponding \chumpy implementation, by leveraging the computing power of modern GPUs.
Examples of this \smplifyPP method are shown in Fig.~\ref{fig:emotionsGETTY}.

To evaluate the accuracy, we need new data with full-body RGB images and corresponding \threeD ground truth bodies.
To that end, we curate a new evaluation dataset containing images of a subject performing a wide variety of poses, gestures and expressions.
We capture \threeD body shape using a scanning system and we fit the \smplHF model to the scans.
This form of pseudo \gt is accurate enough to enable quantitative evaluations for models of body, hands and faces together. 
We find that our model and method performs significantly better than related and less powerful models, resulting in natural and expressive results. 

We believe that this work is a significant step towards \emph{expressive} capture of bodies, hands and faces \emph{together} from a single RGB image. 
We make available for research purposes the \smplHF model, \smplifyPP code, trained networks, model fits, and the evaluation dataset at \websiteSMPLX.

\section{Related work}									\label{sec:related}
\subsection{Modeling the body}

{\bf Bodies, Faces and Hands.}
The problem of modeling the \threeD body has previously been tackled by breaking the body into parts and modeling these parts separately.
We focus on methods that learn statistical shape models from \threeD scans.

Blanz and Vetter \cite{BlanzVetter1999} pioneered this direction with their \threeD morphable face model. 
Numerous methods since then have learned \threeD face shape and expression from scan data; see \cite{Brunton2014_Review,zollhofer2018stateFaces} for recent reviews.
A key feature of such models is that they can represent different face shapes and a wide range of expressions, typically using blend shapes inspired by \facs \cite{EkmanFriesen1978}. 
Most approaches focus only on the face region and not the whole head. 
\flame \cite{li2017learning}, in contrast, models the whole head, captures \threeD head rotations, and also models the neck region; we find this critical for connecting the head and the body. 
None of these methods, model correlations in face shape and body shape. 

The availability of \threeD body scanners enabled learning of body shape from scans.  
In particular the \caesar dataset \cite{CAESAR} opened up the learning of shape \cite{allen2003space}. 
Most early work focuses on body shape using scans of people in roughly the same pose. 
Anguelov et al. \cite{Anguelov05} combined shape with scans of one subject in many poses to learn a factored model of body shape and pose based on triangle deformations. 
Many models followed this, either using triangle deformations \cite{tenbo,Freifeld:ECCV:2012,hasler2009statistical,Hirshberg:ECCV:2012,Dyna:SIGGRAPH:2015} or vertex-based displacements \cite{Allen:2006:LCM,Hasler10,SMPL:2015}, however they all focus on modeling body shape and pose without the hands or face. 
These methods assume that the hand is either in a fist or an open pose and that the face is in a neutral expression. 

Similarly, hand modeling approaches typically ignore the body. 
Additionally,  \threeD hand models  are typically not learned but either are artist designed \cite{srinath_iccv2013}, 
based on shape primitives \cite{Melax:2013:handPhysics,OikonomidisBMVC,DART-Schmidt-RSS-14}, 
reconstructed with multiview stereo and have fixed shape \cite{LucaHands,Tzionas:IJCV:2016}, 
use non-learned per-part scaling parameters \cite{ParagiosHandMonocular2011}, 
or use simple shape spaces \cite{Tkach:SIGGRAPH:2016}. 
Only recently \cite{MSR_2015_CVPR_learnShapeModel,romero2017embodied} have learned  hand models appeared in the literature. 
Khamis \etal \cite{MSR_2015_CVPR_learnShapeModel} collect partial depth maps of $50$ people to learn a model of shape variation, however they do not capture a pose space. 
Romero \etal \cite{romero2017embodied} on the other side learn a parametric hand model (\mano) with both a rich shape and pose space using \threeD scans of $31$ subjects in up to $51$ poses, following the \smpl \cite{SMPL:2015} formulation. 

{\bf Unified Models.}
The most similar models to ours are Frank \cite{joo2018total} and \smplH \cite{romero2017embodied}. 
Frank stitches together three different models: \smpl (with no pose blend shapes) for the body, an artist-created rig for the hands, and the \facewarehouse model \cite{Cao2014_FaceWarehouse} for the face. 
The resulting model is not fully realistic. 
\smplH combines the \smpl body with a \threeD hand model that is learned from \threeD scans. 
The shape variation of the hand comes from full body scans, while the pose dependent deformations are learned from a dataset of hand scans. 
\smplH does not contain a deformable face. 

We start from the publicly-available \smplH \cite{manoWEB} and add the publicly-available \flame head model \cite{flameWEB} to it. 
Unlike Frank, however, we do not simply graft this onto the body. 
Instead we take the full model and fit it to $5586$ \threeD scans and learn the shape and pose-dependent blend shapes. 
This results in a natural looking model with a consistent parameterization. 
Being based on \smpl, it is differentiable and easy to swap into applications that already use \smpl.

\subsection{Inferring the body}

There are many methods that estimate \threeD faces  from images or RGB-D \cite{zollhofer2018stateFaces} as well as methods that estimate hands from such data \cite{hands:Survey:2017}.
While there are numerous methods that estimate the location of \threeD joints from a single image, here we focus on methods that extract a full \threeD body mesh. 

Several methods estimate the \smpl model from a single image \cite{kanazawa2018end,lassner2017unite,omran2018neural,pavlakos2018learning}. 
This is not trivial due to a paucity of training images with paired \threeD model parameters.
To address this, \smplify \cite{bogo2016keep} detects \twoD image features ``bottom up'' and then fits the \smpl model to these ``top down'' in an optimization framework.
In \cite{lassner2017unite} these \smplify fits are used to iteratively curate a training set of paired data to train a direct regression method.
HMR \cite{kanazawa2018end} trains a model without paired data by using \twoD keypoints and an adversary that knows about \threeD bodies.
Like \smplify, NBF \cite{omran2018neural}  uses an intermediate \twoD representation (body part segmentation) and infers \threeD pose from this intermediate representation.
MonoPerfCap \cite{xu2018monoperfcap} infers \threeD pose while also refining surface geometry to capture clothing.
These methods estimate only the \threeD pose of the body without the hands or face.

There are also many multi-camera setups for capturing \threeD pose, \threeD meshes (performance capture), or parametric \threeD models \cite{Ballan3DPVT08,Faugeras_PhysicalForces,gall2009motion,MuVS:3DV:2017,CMU:ICCV:2015,Juergen_NewJournal,Review_Moeslund_2006,rhodin2018learning,hilton2007surface}. 
Most relevant is the  Panoptic studio \cite{CMU:ICCV:2015} which shares our goal of capturing rich, expressive, human interactions.
In \cite{joo2018total}, the \frank model parameters are estimated from multi-camera data by fitting the model to \threeD keypoints and \threeD point clouds.
The capture environment is complex, using $140$ VGA cameras for the body, 
$480$ VGA cameras for the feet, and $31$ HD cameras for the face and hand keypoints.  
We aim for a similar level of expressive detail but from a {\em single RGB image}.

\section{Technical approach}								\label{sec:technical}
In the following we describe \smplHF (Section \ref{sec:smplHF_technical}), and our approach (Section \ref{sec:technical_SMPLifyPP}) for fitting \smplHF to single RGB images. 
Compared to \smplify \cite{bogo2016keep}, \smplifyPP uses 
a better pose prior (Section \ref{sec:technical_VPoser}), 
a more detailed collision penalty (Section \ref{sec:technical_Collisions}),
gender detection (Section \ref{sec:technical_homogenus}), and 
a faster \pytorch implementation (Section \ref{sec:technical_optimization}). 

\subsection{Unified model: \smplHF}	\label{sec:smplHF_technical}

We create a unified model, called \emph{\smplHF}, for \emph{SMPL eXpressive}, with {shape parameters trained jointly for the face, hands and body}. 
\smplHF uses standard vertex-based linear blend skinning with learned corrective blend shapes, has $N=10,475$ vertices and $K=54$ joints, which includes joints for the neck, jaw, eyeballs and fingers. 
\smplHF is defined by a function $M\left( \theta, \beta, \psi \right):\mathbb{R}^{| \theta | \times | \beta | \times | \psi |} \rightarrow \mathbb{R}^{3N}$, parameterized by the pose $\theta \in \mathbb{R}^{3(K+1)} $ where $K$ is the number of body joints in addition to a  joint for global rotation. 
We decompose the pose parameters $\theta$ into: 
$\theta_f$ for the jaw joint, 
$\theta_h$ for the finger joints, and 
$\theta_b$ for the remaining body joints. 
The {joint body, face and hands} shape parameters are noted by $\beta \in \mathbb{R}^{|\beta|} $ and the facial expression parameters by  $\psi \in \mathbb{R}^{|\psi|}$. 
More formally: 
\begin{align}
M   \left(\beta,\theta,\psi\right) &= W \left(T_p \left(\beta,\theta,\psi \right) , J\left(\beta \right),  \theta,\mathcal{W} \right) 									\label{skinning}			\\
T_P \left(\beta,\theta,\psi\right) &= \bar{T} + B_S \left(\beta;\mathcal{S} \right) + B_E  \left( \psi;\mathcal{E} \right) + B_P \left(\theta;\mathcal{P} \right)		\label{smpl_offsets}
\end{align}
where
$B_S \left(\beta;\mathcal{S} \right) = \sum_{n=1}^{|\beta|} \beta_n \mathcal{S}_n$ is the shape blend shape function,  
$\beta$ are linear shape coefficients, $|\beta|$ is their number, 
$\mathcal{S}_n \in \mathbb{R}^{3N}$ are orthonormal principle components of vertex displacements capturing shape variations due to different person identity, and 
$\mathcal{S} = \left [ S_1, \dots , S_{|\beta|} \right ] \in \mathbb{R}^{3N \times |\beta|}$ is a matrix of all such displacements.  
$ B_P \left(\theta;\mathcal{P} \right):\mathbb{R}^{|\theta|} \rightarrow \mathbb{R}^{3N}$ is the pose blend shape function, which adds corrective vertex displacements to the template mesh $\bar{T}$ 
as in \smpl \cite{lopermahmoodetal2014}: 
\begin{equation}
	B_P\left(\theta;\mathcal{P} \right) = \sum_{n=1}^{9K} (R_n(\theta) - R_n(\theta^*) ) \mathcal{P}_n,
\end{equation}
where $R: \mathbb{R}^{|\theta|} \rightarrow \mathbb{R}^{9K}$ is a function mapping the pose vector $\theta$ to a vector of concatenated part-relative rotation matrices, computed with the Rodrigues formula \cite{Malik_Twist,Murray_MathInRob,PonsModelBased} and 
$R_n(\theta)$ is the $n^{th}$ element of $R(\theta)$, 
$\theta^*$ is the pose vector of the rest pose, 
$\mathcal{P}_n \in \mathbb{R}^{3N}$ are again orthonormal principle components of vertex displacements, 
and $\mathcal{P} = \left [ P_1, \dots , P_{9K} \right ] \in \mathbb{R}^{3N \times 9K}$ is a matrix of all pose blend shapes. 
$ B_E \left(\psi;\mathcal{E} \right) =\sum_{n=1}^{|\psi|}\psi_n \mathcal{E}$ is the expression blend shape function, where $\mathcal{E}$ are principle components capturing variations due to facial expressions and $\psi$ are PCA coefficients. 
Since \threeD joint locations $J$ vary between bodies of different shapes, they are a function of body shape $J(\beta) = \mathcal{J} \left( \bar{T} +  B_S \left(\beta;\mathcal{S} \right) \right) $, where $\mathcal{J}$ is a sparse linear regressor that regresses \threeD joint locations from mesh vertices. 
A standard linear blend skinning function $W(.)$ \cite{LBS_PoseSpace} rotates the vertices in $T_p \left(.\right)$ around the estimated joints $J(\beta)$ smoothed by blend weights $\mathcal{W} \in \mathbb{R}^{N \times K}$. 

{We start} with an artist designed \threeD template, whose face and hands match the templates of FLAME \cite{li2017learning} and \mano \cite{romero2017embodied}. 
We fit the template to four datasets of \threeD human scans to get \threeD alignments as training data for \smplHF. 
The shape space parameters, $\left \{ \mathcal{S} \right \}$, are trained on $3800$ alignments in an A-pose capturing variations across identities~\cite{CAESAR}.
The body pose space parameters, $\left \{ \mathcal{W},\mathcal{P},\mathcal{J} \right \}$, are trained on $1786$ alignments in diverse poses.
Since the full body scans have limited resolution for the hands and face, we leverage the  parameters of \mano \cite{romero2017embodied} and \flame \cite{li2017learning}, learned from $1500$ hand and $3800$ head high resolution scans respectively. 
More specifically, we use the pose space and pose corrective blendshapes of \mano for the hands and the expression space $\mathcal{E}$ of \flame. 

The fingers have $30$ joints, which correspond to $90$ pose parameters ($3$ DoF per joint as axis-angle rotations). 
\smplHF uses a lower dimensional PCA pose space for the hands such that $\theta_h = \sum_{n=1}^{|m_h|}m_{h_n} \mathcal{M}$, where $\mathcal{M}$ are principle components capturing the finger pose variations and $m_h$ are the corresponding PCA coefficients. 
{As noted above}, we use the PCA pose space of \mano, that is trained on a large dataset of \threeD articulated human hands. 
The total number of model parameters in \smplHF is $119$: $75$ for the global body rotation and \{ body, eyes , jaw \} joints, $24$ parameters for the lower dimensional hand pose PCA space, $10$ for subject shape and $10$ for the facial expressions. 
Additionally there are separate male and female models, which are used when the gender is known, and a shape space constructed from both genders for when gender is unknown.
\smplHF is realistic, expressive, differentiable and easy to fit to data.

\subsection{\smplifyPP: \smplHF from a single image}						\label{sec:technical_SMPLifyPP}
To fit \smplHF to single RGB images (\emph{\smplifyPP}), we follow \smplify \cite{bogo2016keep} but improve every aspect of it.
We formulate fitting \smplHF to the image as an optimization problem, where we seek to minimize the objective function
\begin{flalign}
  E(\beta,\theta,\psi) = ~
  														&
                                                                 						E_J 					+	\lambda_{\theta_b} 		E_{\theta_b}					+
  															\lambda_{\theta_f}	 	E_{\theta_f}			+
  															\lambda_{m_h}			E_{m_h}        		+ 	\nonumber\\
  		  												&	\lambda_{\alpha}		 	E_{\alpha}			+
  															\lambda_{\beta}    		E_{\beta}			+
  															\lambda_{\mExpr}    		E_{\mExpr}			+
  															\lambda_{\mColl}    		E_{\mColl}
  \label{eq:objective}
\end{flalign}
where $\theta_b$, $\theta_f$ and $m_h$ are the pose vectors for the body, face and the two hands respectively, and $\theta$ is the full set of optimizable pose parameters. 
{The body pose parameters are a function $\theta_b(Z)$, where $Z \in \mathbb{R}^{32}$ is a lower-dimensional pose space described in Section \ref{sec:technical_VPoser}.} 
$E_J(\beta,\theta,K,\Jest)$ is the data term as described below, while the terms $E_{m_h}(m_h)$, $E_{\theta_f}(\theta_f)$, $E_{\beta}(\beta)$ and $E_{\mExpr}(\psi)$ are simple $L_2$ priors for the hand pose, facial pose, body shape and facial expressions, penalizing deviation from the neutral state.
Since the shape space of \smplHF is scaled for unit variance, similarly to \cite{romero2017embodied}, $E_{\beta}(\beta) = \| \beta \|^2$ describes the Mahalanobis distance between the shape parameters being optimized and the shape distribution in the training dataset of \smplHF.
$E_{\alpha}(\theta_b) = \sum_{i \in (elbows,knees)}\exp(\theta_i)$ follows \cite{bogo2016keep} and is a simple prior penalizing extreme bending only for elbows and knees.
We further employ $E_{\theta_b}(\theta_b)$ that is a VAE-based body pose prior (Section \ref{sec:technical_VPoser}), while $E_{\mColl}(\theta_{b,h,f},\beta)$ is an interpenetration penalty (Section \ref{sec:technical_Collisions}).
Finally, $\lambda$ denotes weights that steer the influence of each term in Eq. \ref{eq:objective}.
We empirically find that an annealing scheme for $\lambda$ helps optimization (Section \ref{sec:technical_optimization}).

For the {\it data term} we use a re-projection loss to minimize the weighted robust distance between estimated \twoD joints $\Jest$ and the \twoD projection of the corresponding posed \threeD joints $R_{\theta}( J(\beta))_i$
of \smplHF for each joint $i$, where $R_{\theta}(\cdot)$ is a function that transforms the joints along the kinematic tree according to the pose $\theta$.
Following the notation of \cite{bogo2016keep}, the data term is $   E_J(\beta,\theta,K,\Jest) =$
\begin{equation}
\sum_{joint~i} \gamma_i \omega_i \rho (\mathit{\Pi}_K(  R_{\theta}( J(\beta))_i)    - J_{est,i})
  \label{eq:data_term}
\end{equation}
where $\mathit{\Pi}_K$ denotes the \threeD to \twoD projection with intrinsic camera parameters $K$.
For the \twoD detections we rely on the OpenPose library~\cite{cao2017realtime,simon2017hand,wei2016convolutional}, which provides body, hands, face and feet keypoints jointly for each person in an image.
To account for noise in the detections, the contribution of each joint in the data term is weighted by the detection confidence score $\omega_i$, while $\gamma_i$ are per-joint weights for annealed optimization, as described in Section \ref{sec:technical_optimization}.
Finally, $\rho$ denotes a robust \mbox{Geman-McClure} error function \cite{GemanMcClure1987} for down weighting noisy detections.

\subsection{Variational Human Body Pose Prior}							\label{sec:technical_VPoser}
We seek a prior over body pose that penalizes impossible poses while allowing possible ones. \smplify uses an approximation to the negative log of a Gaussian mixture model trained on \mocap data. 
While effective, we find that the \smplify prior is not sufficiently strong. 
Consequently, we train our body pose prior, \vposer, using a variational autoencoder \cite{kingmawelling2013}, which learns a latent representation of human pose and regularizes the distribution of the latent code to be a normal distribution. 
To train our prior, we use \citeMOSH to recover body pose parameters from three publicly available human motion capture datasets: 
CMU \cite{cmuWEB}, training set of Human3.6M \cite{ionescupapavaetal2014}, and the PosePrior dataset \cite{akhterblack2015}. 
Our training and test data respectively consist of roughly $1$M, and $65$k poses, in rotation matrix representation. 
Details on the data preparation procedure is given in \supmat

The training loss of the VAE is formulated as:
\begin{gather}
\mathcal{L}_{total} = c_1\mathcal{L}_{KL} + c_2\mathcal{L}_{rec} + c_3\mathcal{L}_{orth} + c_4\mathcal{L}_{det1} + c_5\mathcal{L}_{reg} \label{eq:loss_total}\\
	\mathcal{L}_{KL} = KL(q(Z|R)||\mathcal{N}(0,I))\label{eq:loss_kl}\\
	\mathcal{L}_{rec} = ||R - \hat{R}||_2^{2}\label{eq:loss_rec}\\
	\mathcal{L}_{orth} = ||\hat{R}\hat{R}^{'} - I||_2^{2}\label{eq:loss_orth}\\
	\mathcal{L}_{det1} = |det(\hat{R}) - 1|\label{eq:loss_det1}\\
	\mathcal{L}_{reg} = ||\phi||_2^{2},\label{eq:loss_reg}
\end{gather}
where $Z \in \mathbb{R}^{32}$ is the latent space of the autoencoder, $R \in SO(3)$ are $3\times3$ rotation matrices for each joint as the network input and $\hat{R}$ is a similarly shaped matrix representing the output. 
The Kullback-Leibler term in Eq. \eqref{eq:loss_kl}, and the reconstruction term in Eq. \eqref{eq:loss_rec} follow the VAE formulation in \cite{kingmawelling2013}, while their role is to encourage a normal distribution on the latent space, and to make an efficient code to reconstruct the input with high fidelity. 
Eq. \eqref{eq:loss_orth} and \eqref{eq:loss_det1} encourage the latent space to encode valid rotation matrices. 
Finally, Eq. \eqref{eq:loss_reg} helps prevent over-fitting by encouraging smaller network weights $\phi$. 
Implementation details can be found in \supmat

To employ \vposer in the optimization, rather than to optimize over $\theta_b$ directly in Eq.~\ref{eq:objective}, we optimize the parameters of a $32$ dimensional latent space with a quadratic penalty on $Z$ and transform this back into joint angles $\theta_b$ in axis-angle representation. 
This is analogous to how hands are treated except that the hand pose $\theta_h$ is projected into a linear PCA space and the penalty is on the linear coefficients.

\subsection{Collision penalizer}											\label{sec:technical_Collisions}
When fitting a model to observations, there are often self-collisions and penetrations of several body parts that are physically impossible.
Our approach is inspired by \smplify, that penalizes penetrations with an underlying collision model based on shape primitives, \ie an ensemble of capsules. 
Although this model is computationally efficient, it is only a rough approximation of the human body. 

For models like \smplHF, that also model the fingers and facial details, a more accurate collision model in needed. 
To that end, we employ the detailed collision-based model for meshes from \cite{LucaHands,Tzionas:IJCV:2016}.
We first detect a list of colliding triangles $\mathcal{C}$ by employing Bounding Volume Hierarchies (BVH) \cite{collisionDeformableObjects} and compute local conic \threeD distance fields $\Psi$ defined by the triangles $\mathcal{C}$ and their normals $\normal$. 
Penetrations are then penalized by the depth of intrusion, efficiently computed by the position in the distance field. 
For two colliding triangles $f_s$ and $f_t$, intrusion is bi-directional; 
the vertices $\vertex_t$ of $f_t$ are the \emph{intruders} in the distance field $\Psi_{f_s}$ of the \emph{receiver} triangle $f_s$ and are penalized by $\Psi_{f_s}(\vertex_t)$, and vice-versa. 
Thus, the collision term $E_{\mColl}$ in the objective (Eq. \ref{eq:objective}) is defined as
\begin{equation}\label{eq:collisionDetection_p2p}
\begin{split}
	E_{\mColl}(\para) = \sum_{(f_s(\para), f_t(\para))\in \mathcal{C}} \biggl\{ &\sum_{\vertex_s \in f_s} \Vert -\Psi_{f_t}(\vertex_s)\normal_s  \Vert^2 + \\ &\sum_{\vertex_t \in f_t} \Vert -\Psi_{f_s}(\vertex_t)\normal_t \Vert^2 \biggr\} .
	\end{split}
\end{equation}
For technical details about $\Psi$, as well as details about handling collisions for parts with permanent or frequent self-contact we redirect the reader to \cite{LucaHands,Tzionas:IJCV:2016} and \supmat.
For computational efficiency, we use a highly parallelized implementation of BVH following \cite{Karras:2012:MPC} with a custom \cuda kernel wrapped around a custom \pytorch operator.

\subsection{Deep Gender Classifier}										\label{sec:technical_homogenus}
Men and women have different proportions and shapes. Consequently, using the appropriate body model to fit \twoD data means that we should apply the appropriate shape space. 
We know of no previous method that automatically takes gender into account in fitting \threeD human pose. 
In this work, we train a gender classifier that takes as input an image containing the full body and the \openpose joints, and assigns a gender label to the detected person. 
To this end, we first annotate through Amazon Mechanical Turk a large dataset of images from LSP \cite{johnson2010clustered}, LSP-extended \cite{johnson2011learning}, MPII \cite{andriluka2014mpii}, MS-COCO\cite{linmaireetal2014}, and LIP datset \cite{liangxuetal2017}, while following their official splits for train and test sets. 
The final dataset includes 50216 training examples and 16170 test samples (see \supmat). 
We use this dataset to fine tune a pretrained ResNet18 \cite{hezhangetal2015a} for binary gender classification. 
Moreover, we threshold the computed class probabilities, by using a class-equalized validation set, to obtain a good trade-off between discarded, correct, and incorrect predictions. 
We choose a threshold of 0.9 for accepting a predicted class, which yields $62.38\%$ correct predictions, and $7.54\%$ incorrect predictions on the validation set.
At test time, we run the detector and fit the appropriate gendered model. 
When the detected class probability is below the threshold, we fit the gender-neutral body model.

\subsection{Optimization}												\label{sec:technical_optimization}
\smplify employs \chumpy and \openDR \cite{opendr} which makes the optimization slow.
To keep optimization of Eq.~\ref{eq:objective} tractable,  we use \pytorch and the Limited-memory BFGS optimizer (L-BFGS) \cite{nocedal2006nonlinear} with strong Wolfe line search. 
Implementation details can be found in \supmat

We optimize  Eq. \ref{eq:objective} with a multistage approach, similar to \cite{bogo2016keep}. 
We assume that we know the exact or an approximate value for the focal length of the camera. 
Then we first estimate the unknown camera translation and global body orientation (see  \cite{bogo2016keep}).
We then fix the camera parameters and optimize body shape, $\beta$, and pose, $\theta$. 
Empirically, we found that an \emph{annealing scheme} for the weights $\gamma$ in the data term $E_J$ (Eq.~\ref{eq:data_term}) helps optimization of the objective (Eq.~\ref{eq:objective}) to deal with ambiguities and local optima. 
This is mainly motivated by the fact that small body parts like the hands and face have many keypoints relative to their size, and can dominate in Eq.~\ref{eq:objective}, throwing optimization in a local optimum when the initial estimate is away from the solution. 

In the following, we denote by $\gamma_b$ the weights corresponding to the main body keypoints, $\gamma_h$ the ones for hands and $\gamma_f$ the ones for facial keypoints. 
We then follow three steps, starting with high regularization to mainly refine the global body pose, 
and gradually increase the influence of hand keypoints to refine the pose of the arms. 
After converging to a better pose estimate, we increase the influence of both hands and facial keypoints to capture expressivity. 
Throughout the above steps the weights $\lambda_{\alpha},\lambda_{\beta},\lambda_{\mExpr}$ in Eq.\ref{eq:objective} start with high regularization that gradually lowers to allow for better fitting, 
The only exception is $\lambda_{\mColl}$ that gradually increases while the influence of hands gets stronger in $E_J$ and more collisions are expected.

\section{Experiments}									\label{sec:experiments}
\subsection{Evaluation datasets}\label{sec:experimentsDatasets}

Despite the recent interest in more expressive models \cite{joo2018total,romero2017embodied} there exists no dataset containing images with \gt shape for bodies, hands and faces together. 
Consequently, we create a dataset for evaluation from currently available data through fitting and careful curation.

\textbf{Expressive hands and faces dataset (EHF).}
We begin with the \smplH dataset \cite{manoWEB}, obtaining one full body RGB image per frame. 
We then align \smplHF to the 4D scans following \cite{romero2017embodied}. 
An expert annotator manually curated the dataset to select $100$ frames that can be confidently considered pseudo \gt, according to alignment quality and interesting hand poses and facial expressions. 
The pseudo \gt meshes allow to use a stricter \emph{vertex-to-vertex (v2v)} error metric~\cite{SMPL:2015,pavlakos2018learning}, in contrast to the common paradigm of reporting \threeD joint error, which does not capture surface errors and rotations along the bones. 

\begin{table}
	\centering
	\hspace{-3mm}
	\tabcolsep=0.75mm
	\begin{tabular}{@{}cccc@{}}
	\toprule
	Model		& Keypoints			& v2v error		& Joint error	\\
	\midrule
	``\smpl''	& Body				& $57.6$			& $63.5$			\\
	``\smpl''	& Body+Hands+Face	& $64.5$			& $71.7$			\\
	``\smplH''	& Body+Hands   		& $54.2$			& $63.9$			\\
	\smplHF		& Body+Hands+Face	& \bf{$52.9$}	& \bf{$62.6$}	\\
	\bottomrule
	\end{tabular}
	\vspace{-2mm}
	\caption{
						Quantitative comparison of ``\smpl'', ``\smplH'' and \smplHF, as described in Section \ref{sec:experimentsEvaluation}, fitted with \smplifyPP on the EHF dataset. 
						We report the mean vertex-to-vertex (v2v) and the standard mean \threeD body (only) joint error in mm. 
						The table shows that richer modeling power results in lower errors. 
	}
	\label{tab:SIGA_dataset}
\end{table}
 
\begin{table}
\centering
	\begin{tabular}{l|c}
	\toprule
	Version & v2v error\\
	\midrule
	\smplifyPP 								& $52.9$ \\
	\quad		gender neutral model 		& $58.0$ \\
	\quad		replace Vposer with GMM		& $56.4$ \\
	\quad        no collision term 			& $53.5$ \\
	\bottomrule
	\end{tabular}
	\vspace*{-2mm}
	\caption{					Ablative study for \smplifyPP on the EHF dataset. 
								The numbers reflect the contribution of each component in overall accuracy.
	}
\label{table:SIGA_Ablative}
\end{table}

\subsection{Qualitative \& Quantitative evaluations}\label{sec:experimentsEvaluation}

To test the effectiveness of  \smplHF and \smplifyPP, we perform comparisons to the most related models, namely \smpl \cite{SMPL:2015}, \smplH \cite{romero2017embodied}, and \frank \cite{joo2018total}. 
In this direction we fit \smplHF to the EHF images to evaluate both \emph{qualitatively} and \emph{quantitatively}. 
Note that we use \emph{only} $1$ image and \twoD joints as input, while previous methods use \emph{much more} information; \ie \threeD point clouds \cite{joo2018total,romero2017embodied} and joints \cite{joo2018total}. 
Specifically \cite{SMPL:2015,romero2017embodied} employ $66$ cameras and $34$ projectors, while \cite{joo2018total} employ more than $500$ cameras. 

We first compare to \smpl, \smplH and \smplHF on the EHF dataset and report results in Table \ref{tab:SIGA_dataset}. 
The table reports \emph{mean vertex-to-vertex  (v2v)} error and \emph{mean \threeD body joint} error after Procrustes alignment with the \gt \threeD meshes and body (only) joints respectively. 
To ease numeric evaluation, for this table only we ``simulate'' \smpl and \smplH with a \smplHF variation with locked degrees of freedom, noted as ``\smpl'' and ``\smplH'' respectively. 
As expected, the errors show that the standard mean \threeD joint error fails to capture accurately the difference in model expressivity. 
On the other hand, the much stricter v2v metric shows that enriching the body with finger and face modeling results in lower errors. 
We also fit \smpl with additional features for parts that are not properly modeled, \eg finger features.
The additional features result in an increasing error, pointing to the importance of richer and more expressive models. 
We report similar qualitative comparisons in \supmat

\newcommand{\ImgEmotionsHeightGG}{0.152}
\newcommand{\ImgEmotionsHSpaceGG}{-3.0mm}
\newcommand{\ImgEmotionsVSpaceGG}{-4.5mm}

\begin{figure}[t]
	\centering
	\subfloat{	\includegraphics[trim=000mm 000mm 000mm 02mm, clip=true, width=0.91 \linewidth]{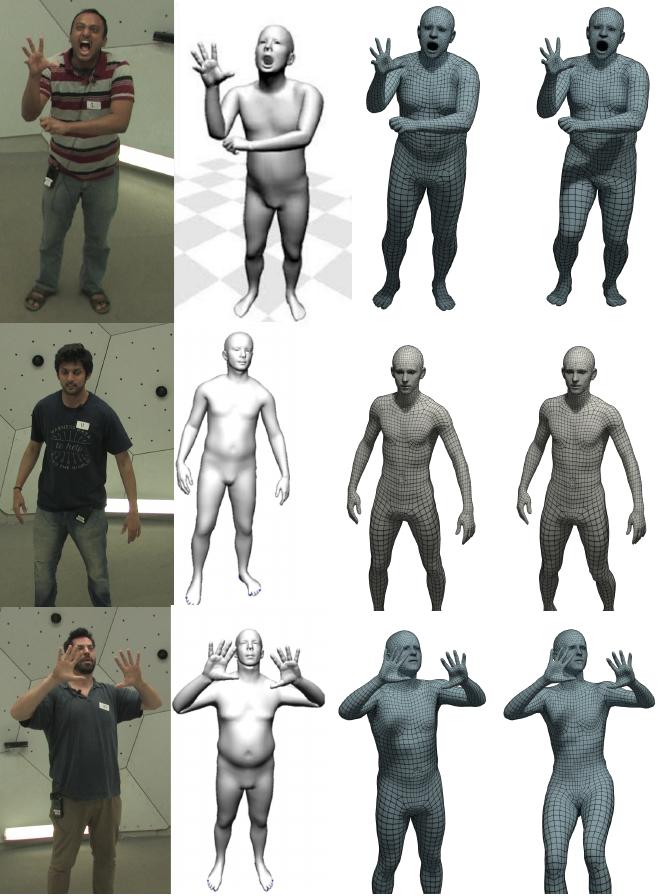}	}
	{
	\small
	\tabcolsep=4.0mm
	\begin{tabular}{cccc}
	\hspace*{+0.5em}		reference	&	\hspace*{-1.0em}		\cite{joo2018total}: $>500$ 		&	\hspace*{-1.0em}		Ours: $>500$		&	\hspace*{-1.0em}		Ours: $1$ 	\\
	\hspace*{+0.5em}		RGB			&	\hspace*{-1.0em}							cameras		&	\hspace*{-1.0em}			  cameras	&	\hspace*{-1.0em}		camera		\\
	\end{tabular}
    }
	\vspace*{-0.5em}
	\caption{						Qualitative comparison of our gender neutral model (top, bottom rows) or gender specific model (middle) against \frank~\cite{joo2018total} on some of their data.
									To fit \frank, \cite{joo2018total} employ both \threeD joints and point cloud, \ie more than $500$ cameras.
									{In contrast, our method produces a realistic and expressive reconstruction using \emph{only} \twoD joints}.
									{
									We show results using the \threeD joints of \cite{joo2018total} projected in $1$ camera view (third column), as well as using joints estimated from only $1$ image (last column),
									to show the influence of noise in \twoD joint detection.
									}
									Compared to \frank, our \smplHF does \emph{not} have skinning artifacts around the joints, \eg elbows.
	}
\label{fig:frank_VS_smplhf}
\vspace*{-1.0em}
\end{figure}

\newcommand{\ImgWildHeightXXX}{0.235}
\newcommand{\ImgWildHSpaceXXX}{-2.0mm}
\newcommand{\ImgWildVSpaceXXX}{-5.0mm}
\begin{figure*}[h!]
	\centering
	\includegraphics[width=0.88 \linewidth,trim={000mm 000mm 000mm 000mm},clip]{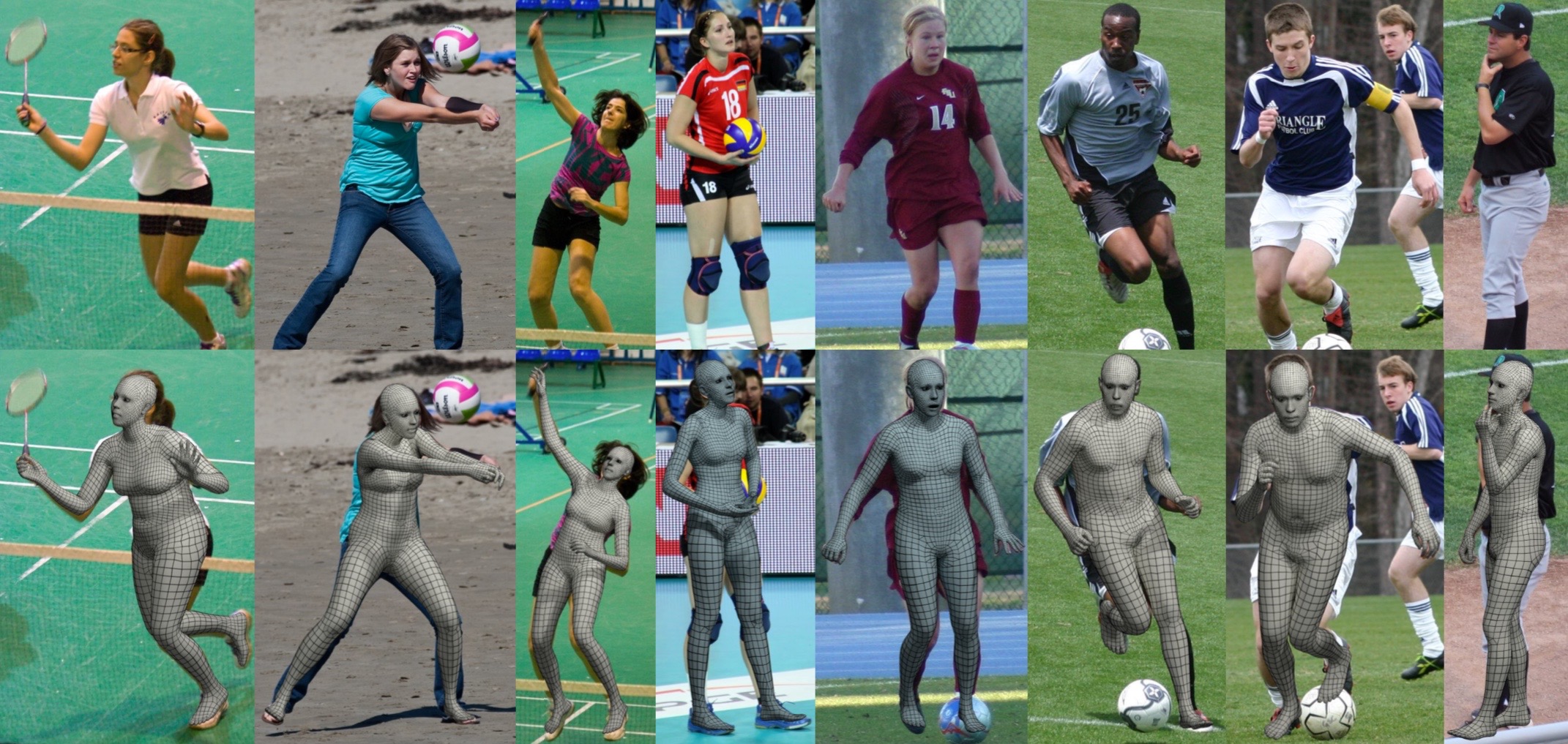}
	\vspace*{-2.0mm}
	\caption{
				Qualitative results of \smplHF for the in-the-wild images of the LSP dataset \cite{johnson2010clustered}. 
				A strong holistic model like \smplHF results in \emph{natural} and \emph{expressive} reconstruction of bodies, hands and faces. %
				Gray color depicts the gender-specific model for confident gender detections. Blue is the gender-neutral model that is used when the gender classifier is uncertain.
	}
\label{fig:inthewild_success}
\vspace*{-3mm}
\end{figure*}

We then perform an ablative study, summarized in Table~\ref{table:SIGA_Ablative}, where we report the \emph{mean vertex-to-vertex (v2v)} error. 
\smplifyPP with a gender-specific model achieves $52.9$ mm error. 
The gender neutral model is easier to use, as it does not need gender detection, but comes with a small compromise in terms of accuracy. 
Replacing \vposer with the GMM of \smplify \cite{bogo2016keep} increases the error to $56.4$ mm, showing the effectiveness of \vposer. 
Finally, removing the collision term increases the error as well, to $53.5$ mm, while also allowing for non physically plausible pose estimates. 

\newcommand{\ImgEmotionsHeightFF}{0.150}
\newcommand{\ImgEmotionsHSpaceFF}{-3.0mm}
\newcommand{\ImgEmotionsVSpaceFF}{-8.0mm}

\begin{figure}[t]
	\vspace*{-2.0mm}
	\centering
	\subfloat{	\includegraphics[trim=150mm 235mm 150mm 30mm,   clip=true, width=\ImgEmotionsHeightFF \textwidth]{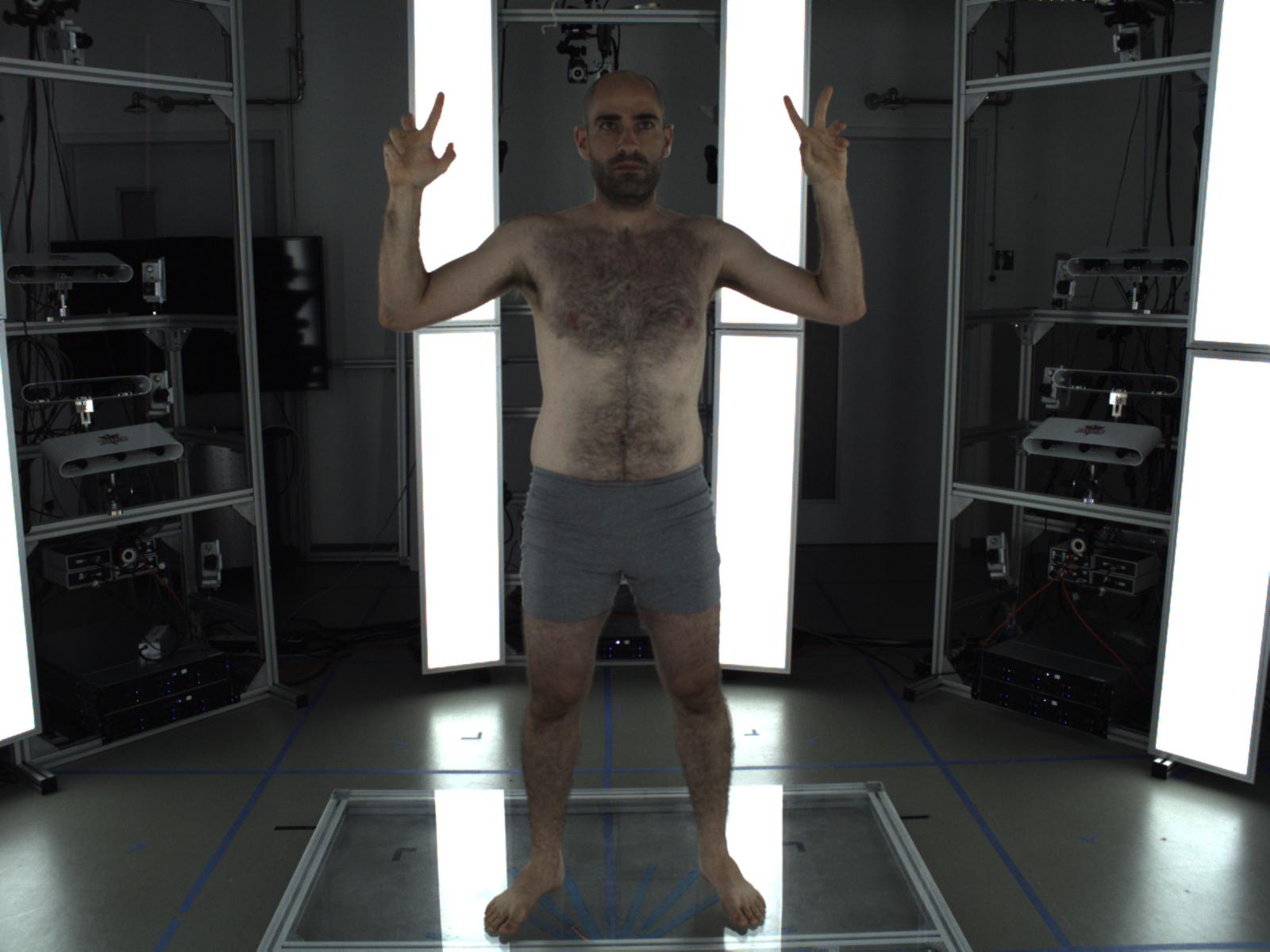}							}		\hspace*{\ImgEmotionsHSpaceFF}
	\subfloat{	\includegraphics[trim=150mm 235mm 150mm 30mm,   clip=true, width=\ImgEmotionsHeightFF \textwidth]{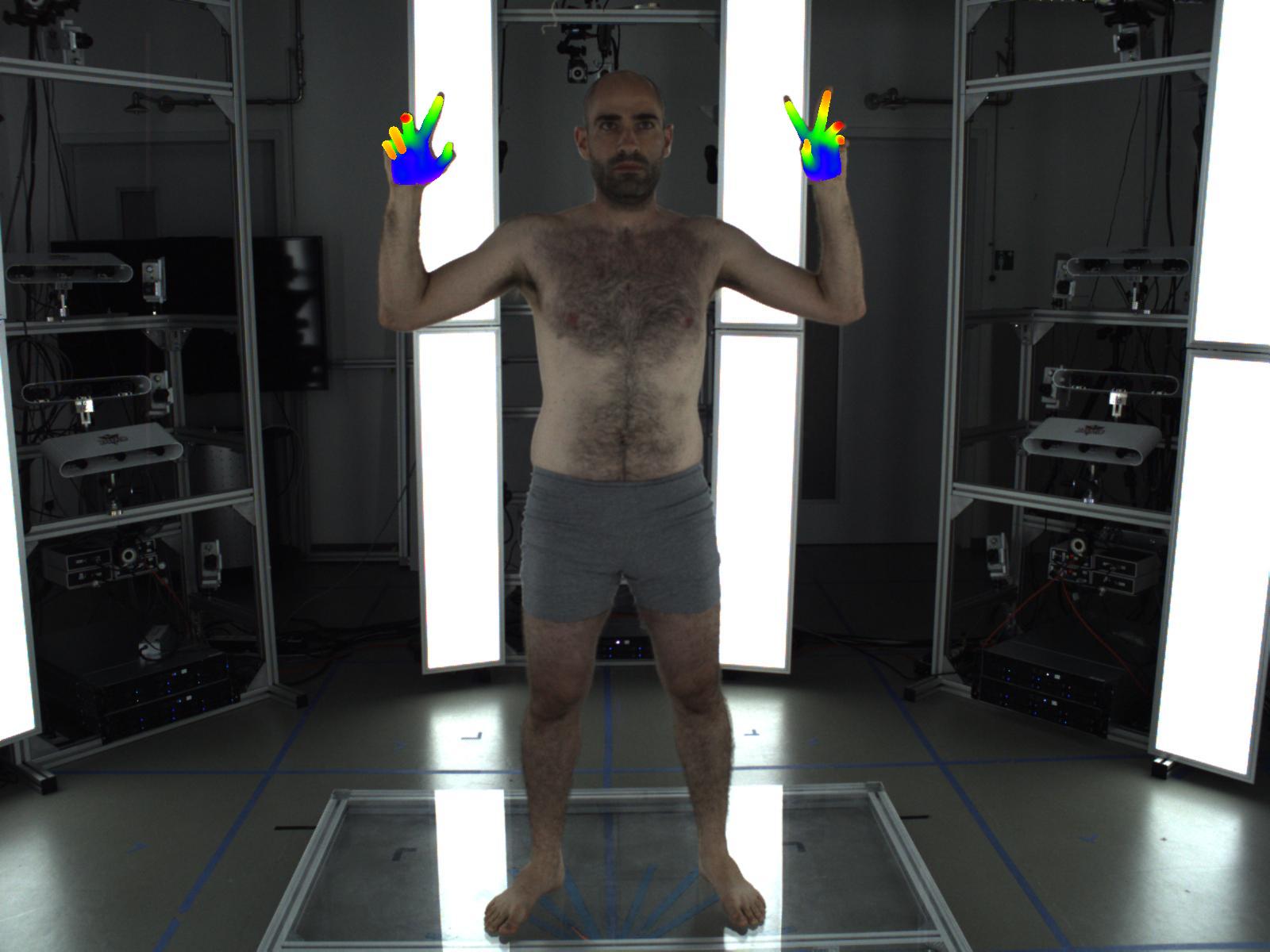}							}		\hspace*{\ImgEmotionsHSpaceFF}
	\subfloat{	\includegraphics[trim=150mm 235mm 150mm 30mm,   clip=true, width=\ImgEmotionsHeightFF \textwidth]{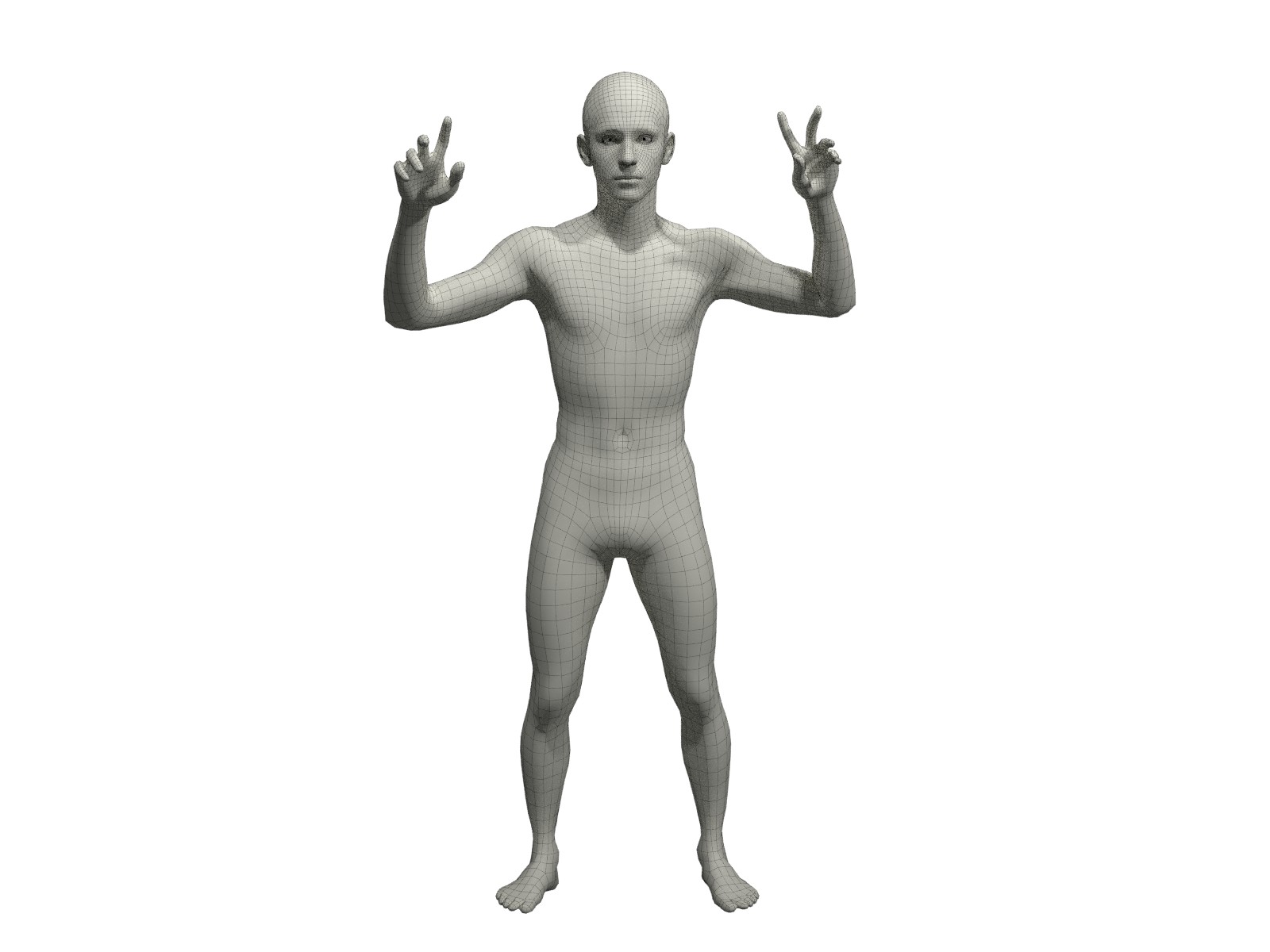}	}	\\	\vspace*{\ImgEmotionsVSpaceFF}
	\subfloat{	\includegraphics[trim=150mm 205mm 150mm 35mm,   clip=true, width=\ImgEmotionsHeightFF \textwidth]{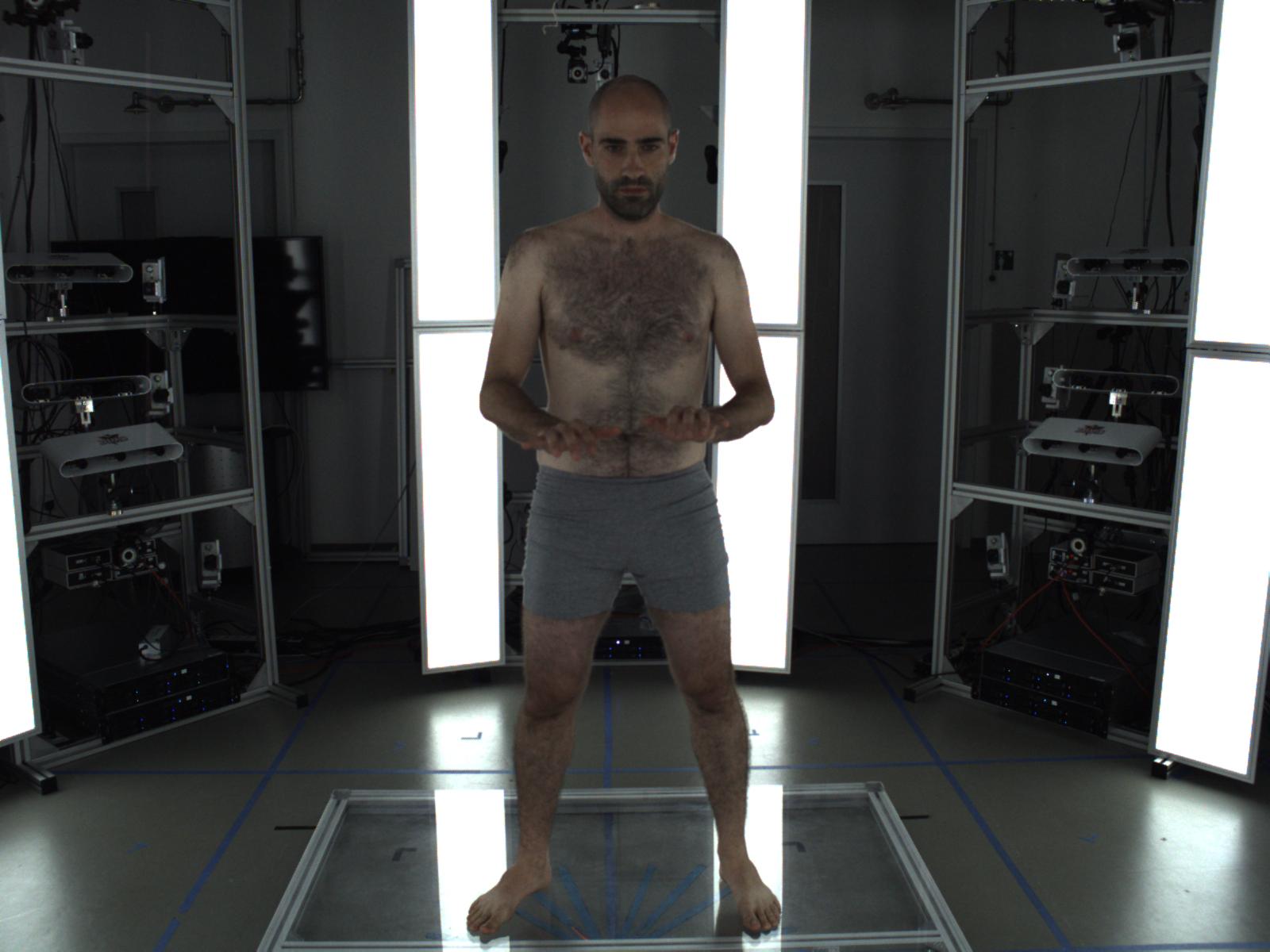}							}		\hspace*{\ImgEmotionsHSpaceFF}
	\subfloat{	\includegraphics[trim=150mm 205mm 150mm 35mm,   clip=true, width=\ImgEmotionsHeightFF \textwidth]{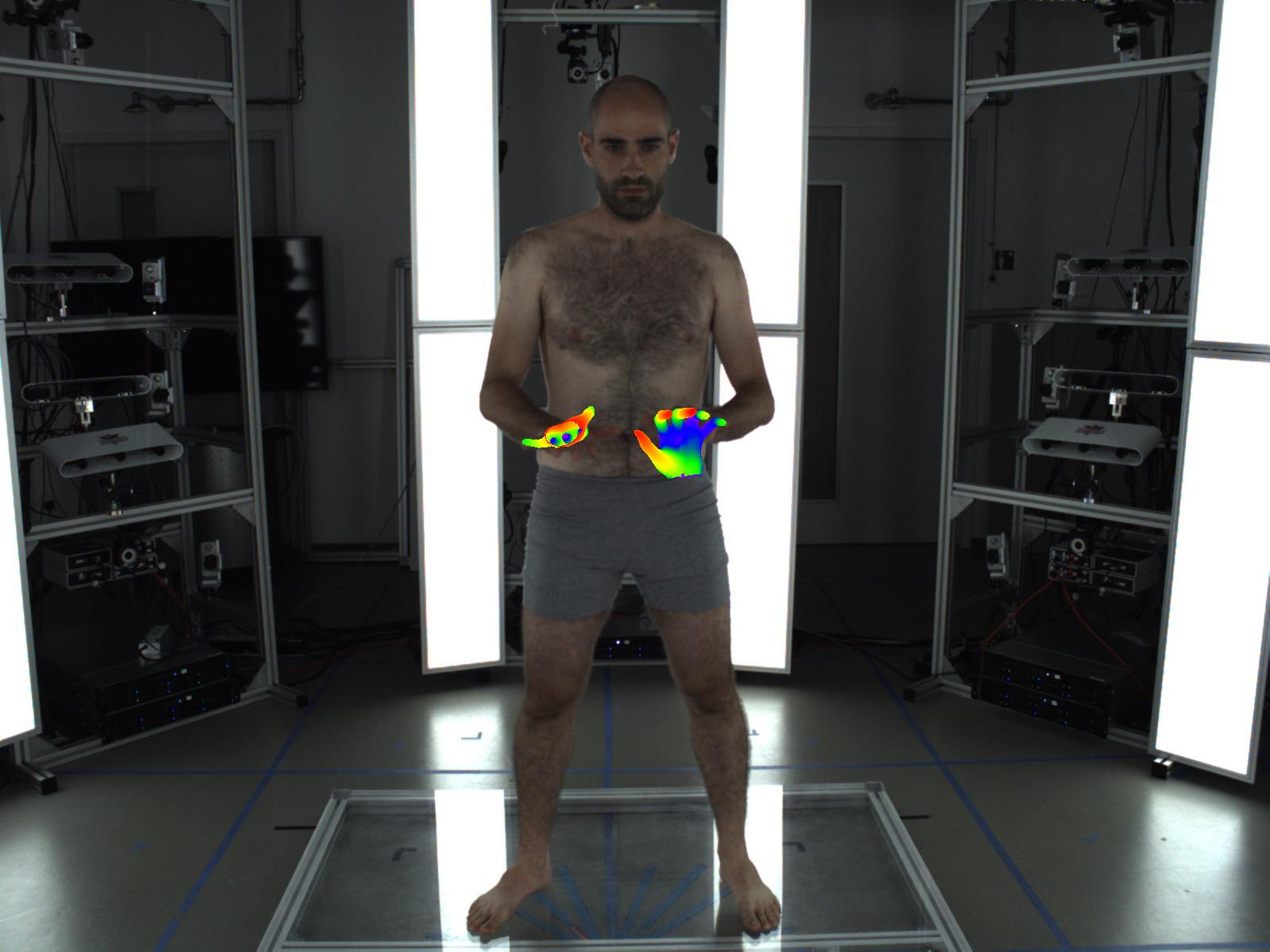}							}		\hspace*{\ImgEmotionsHSpaceFF}
	\subfloat{	\includegraphics[trim=150mm 205mm 150mm 35mm,   clip=true, width=\ImgEmotionsHeightFF \textwidth]{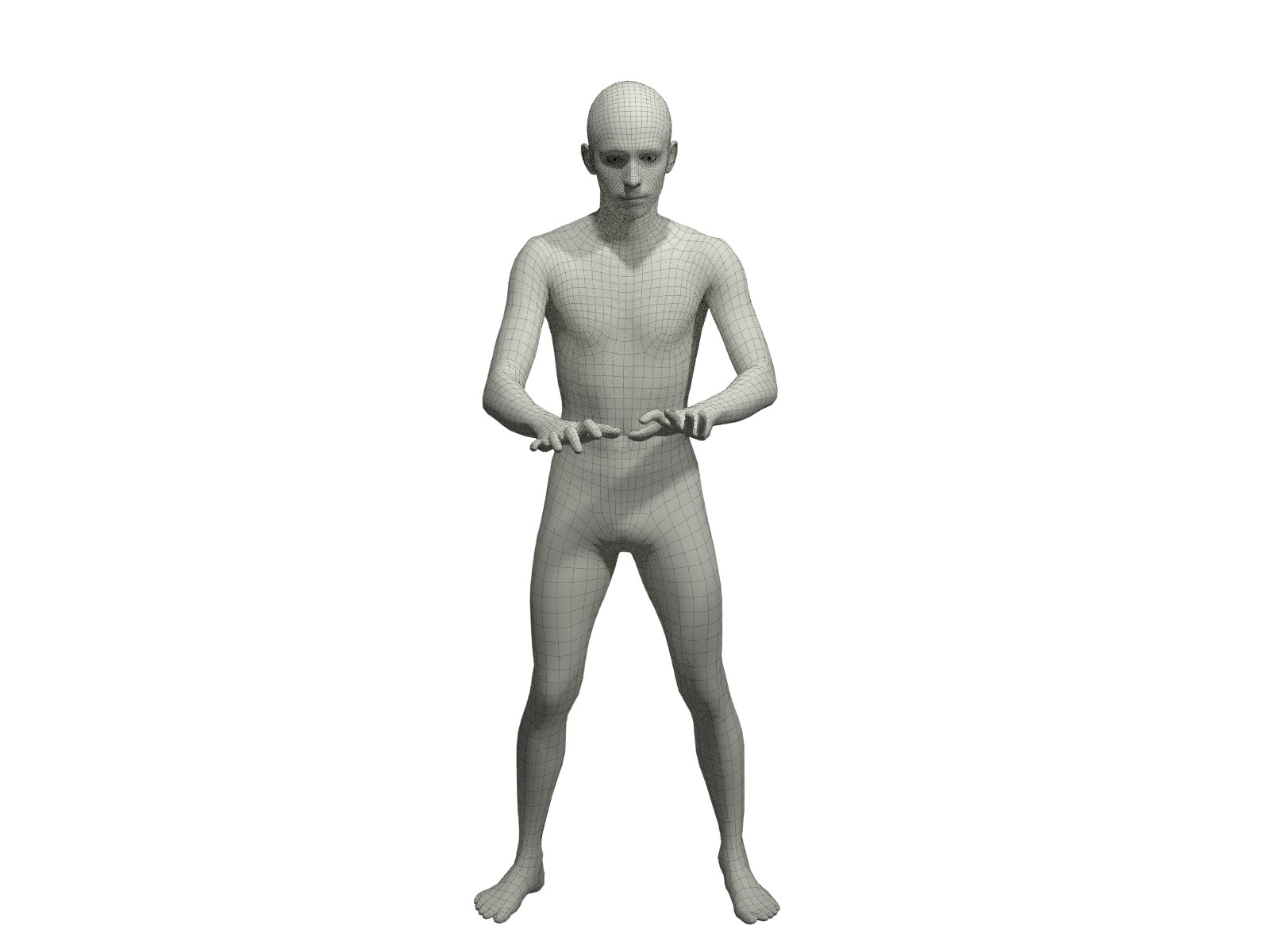}	}
	\vspace*{-2.0mm}
	\caption{
									Comparison of the hands-only approach of \cite{panteleris:WACV:2018} (middle) against our approach with the male model (right). 
									Both approaches depend on \openpose. 
									In case of good detections both perform well (top). 
									In case of noisy \twoD detections (bottom) our holistic model shows increased robustness. 
									(images cropped at the bottom in the interest of space)
	}
\label{fig:hand-only}
\vspace*{-1.5em}
\end{figure}

The closest comparable model to  \smplHF  is \frank \cite{joo2018total}. 
Since \frank is not available to date, nor are the fittings to \cite{TotalCaptureDatasetWEB}, we show images of results found online. 
Figure \ref{fig:frank_VS_smplhf} shows \frank fittings to \threeD joints \emph{and} point clouds, \ie using more than $500$ cameras. 
Compare this with \smplHF fitting that is done with \smplifyPP using \emph{only} $1$ RGB image with \twoD joints. 
For a more direct comparison here, we fit \smplHF to \twoD projections of the \threeD joints that \cite{joo2018total} used for \frank. 
Although we use \emph{much less} data, \smplHF shows at least similar expressivity to \frank for both the face and hands. 
Since \frank does not use pose blend shapes, it suffers from skinning artifacts around the joints, \eg elbows, as clearly seen in Figure \ref{fig:frank_VS_smplhf}. 
\smplHF by contrast, is trained to include pose blend shapes and does not suffer from this. 
As a result it looks more \emph{natural} and \emph{realistic}. 

To further show the value of a holistic model of the body, face and hands, in Fig.~\ref{fig:hand-only} we compare \smplHF and \smplifyPP to the hands-only approach of \cite{panteleris:WACV:2018}. 
Both approaches employ \openpose for \twoD joint detection, while \cite{panteleris:WACV:2018} further depends on a hand detector. 
As seen in Fig.~\ref{fig:hand-only}, in case of good detections both approaches perform nicely, though in case of noisy detections, \smplHF shows increased robustness due to the context of the body. 
We further perform quantitative comparison after aligning the resulting fittings to EHF. 
Due to different mesh topology, for simplicity we use hand joints as pseudo \gt, and perform Procrustes analysis of each hand independently, ignoring the body. 
Panteleris~\etal~\cite{panteleris:WACV:2018} achieve a mean \threeD joint error of $26.5$ mm, while \smplHF has $19.8$ mm.

Finally, we fit \smplHF with \smplifyPP to some \inthewild datasets, namely the LSP \cite{johnson2010clustered}, LSP-extended \cite{johnson2011learning} and MPII datasets \cite{andriluka2014mpii}. 
Figure \ref{fig:inthewild_success} shows some qualitative results for the LSP dataset \cite{johnson2010clustered}; see \supmat for more examples and failure cases.
The images show that a strong holistic model like \smplHF can effectively give \emph{natural} and \emph{expressive} reconstruction from everyday images.

\section{Conclusion}										\label{sec:conclusion}
In this work we present \smplHF, a new model that \emph{jointly} captures the body together with face and hands. 
We additionally present \smplifyPP, an approach to fit \smplHF to a single RGB image and \twoD \openpose joint detections. 
We regularize fitting under ambiguities with a new powerful body pose prior and a fast and accurate method for detecting and penalizing penetrations. 
We present a wide range of qualitative results using images in-the-wild, showing the expressivity of \smplHF and effectiveness of \smplifyPP. 
We introduce a curated dataset with pseudo \gt to perform quantitative evaluation, that shows the importance of more expressive models. 
In future work we will curate a dataset of \inthewild \smplHF fits and learn a regressor to directly regress \smplHF parameters directly from RGB images. 
We believe that this work is an important step towards \emph{expressive} capture of bodies, hands and faces \emph{together} from an RGB image.

\vspace{+01.00mm}
\footnotesize
\noindent
{\bf Acknowledgements:} 
We thank Joachim Tesch for the help with Blender rendering and Pavel Karasik for the help with Amazon Mechanical Turk. 
We thank Soubhik Sanyal for the face-only baseline, 
Panteleris~\etal from FORTH for running their hands-only method~\cite{panteleris:WACV:2018} on the EHF dataset, and 
Joo~\etal from CMU for providing early access to their data \cite{joo2018total}.

\vspace{+01.00mm}
\footnotesize
\noindent
{\bf Disclosure:} 
MJB has received research gift funds from Intel, Nvidia, Adobe, Facebook, and Amazon. 
While MJB is a part-time employee of Amazon, his research was performed solely at, and funded solely by, MPI. 
MJB has financial interests in Amazon and Meshcapade GmbH.

{\small
\balance
\bibliographystyle{ieee_fullname}
\bibliography{egbib,bib_bodies,bib_handrefs,bib_faces,bib_vposer,bib_homogenus}

\begin{thebibliography}{10}\itemsep=-1pt

\bibitem{akhterblack2015}
Ijaz Akhter and Michael~J. Black.
\newblock Pose-conditioned joint angle limits for 3{D} human pose
  reconstruction.
\newblock In {\em CVPR}, 2015.

\bibitem{allen2003space}
Brett Allen, Brian Curless, and Zoran Popovi{\'c}.
\newblock The space of human body shapes: Reconstruction and parameterization
  from range scans.
\newblock {\em ACM Transactions on Graphics, (Proc. SIGGRAPH)}, 22(3):587--594,
  2003.

\bibitem{Allen:2006:LCM}
Brett Allen, Brian Curless, Zoran Popovi\'{c}, and Aaron Hertzmann.
\newblock Learning a correlated model of identity and pose-dependent body shape
  variation for real-time synthesis.
\newblock In {\em ACM SIGGRAPH/Eurographics Symposium on Computer Animation},
  SCA '06, pages 147--156. Eurographics Association, 2006.

\bibitem{Amberg2008}
Brian Amberg, Reinhard Knothe, and Thomas Vetter.
\newblock Expression invariant {3D} face recognition with a morphable model.
\newblock In {\em International Conference on Automatic Face Gesture
  Recognition}, 2008.

\bibitem{andriluka2014mpii}
Mykhaylo Andriluka, Leonid Pishchulin, Peter Gehler, and Bernt Schiele.
\newblock {2D} human pose estimation: New benchmark and state of the art
  analysis.
\newblock In {\em CVPR}, 2014.

\bibitem{Anguelov05}
Dragomir Anguelov, Praveen Srinivasan, Daphne Koller, Sebastian Thrun, Jim
  Rodgers, and James Davis.
\newblock {SCAPE: Shape Completion and Animation of PEople}.
\newblock {\em ACM Transactions on Graphics, (Proc. SIGGRAPH)}, 24(3):408--416,
  2005.

\bibitem{Ballan3DPVT08}
Luca Ballan and Guido~Maria Cortelazzo.
\newblock Marker-less motion capture of skinned models in a four camera set-up
  using optical flow and silhouettes.
\newblock In {\em International Symposium on {3D} Data Processing,
  Visualization and Transmission (3DPVT)}, 2008.

\bibitem{LucaHands}
Luca Ballan, Aparna Taneja, Juergen Gall, Luc Van~Gool, and Marc Pollefeys.
\newblock Motion capture of hands in action using discriminative salient
  points.
\newblock In {\em ECCV}, 2012.

\bibitem{BlanzVetter1999}
Volker Blanz and Thomas Vetter.
\newblock A morphable model for the synthesis of {3D} faces.
\newblock In {\em SIGGRAPH}, pages 187--194, 1999.

\bibitem{bogo2016keep}
Federica Bogo, Angjoo Kanazawa, Christoph Lassner, Peter Gehler, Javier Romero,
  and Michael~J Black.
\newblock Keep it {SMPL}: Automatic estimation of 3{D} human pose and shape
  from a single image.
\newblock In {\em ECCV}, 2016.

\bibitem{Booth2017}
James Booth, Anastasios Roussos, Allan Ponniah, David Dunaway, and Stefanos
  Zafeiriou.
\newblock Large scale 3{D} morphable models.
\newblock {\em IJCV}, 126(2-4):233--254, 2018.

\bibitem{Malik_Twist}
Christoph Bregler, Jitendra Malik, and Katherine Pullen.
\newblock Twist based acquisition and tracking of animal and human kinematics.
\newblock {\em International Journal of Computer Vision (IJCV)},
  56(3):179--194, 2004.

\bibitem{Brunton2014_Review}
Alan Brunton, Augusto Salazar, Timo Bolkart, and Stefanie Wuhrer.
\newblock Review of statistical shape spaces for {3D} data with comparative
  analysis for human faces.
\newblock {\em CVIU}, 128(0):1--17, 2014.

\bibitem{Cao2014_FaceWarehouse}
Chen Cao, Yanlin Weng, Shun Zhou, Yiying Tong, and Kun Zhou.
\newblock Facewarehouse: A 3{D} facial expression database for visual
  computing.
\newblock {\em IEEE Transactions on Visualization and Computer Graphics},
  20(3):413--425, 2014.

\bibitem{cao2017realtime}
Zhe Cao, Tomas Simon, Shih-En Wei, and Yaser Sheikh.
\newblock Realtime multi-person 2{D} pose estimation using part affinity
  fields.
\newblock In {\em CVPR}, 2017.

\bibitem{tenbo}
Yinpeng Chen, Zicheng Liu, and Zhengyou Zhang.
\newblock Tensor-based human body modeling.
\newblock In {\em CVPR}, 2013.

\bibitem{cmuWEB}
CMU.
\newblock {{CMU MoCap dataset}}.

\bibitem{TotalCaptureDatasetWEB}
Total~Capture Dataset.
\newblock \url{http://domedb.perception.cs.cmu.edu}.

\bibitem{ParagiosHandMonocular2011}
Martin {D}e~{L}a Gorce, David~J. Fleet, and Nikos Paragios.
\newblock Model-based {3D} hand pose estimation from monocular video.
\newblock {\em PAMI}, 33(9):1793--1805, 2011.

\bibitem{Faugeras_PhysicalForces}
Quentin Delamarre and Olivier~D. Faugeras.
\newblock {3D} articulated models and multiview tracking with physical forces.
\newblock {\em CVIU}, 81(3):328--357, 2001.

\bibitem{EkmanFriesen1978}
P. Ekman and W. Friesen.
\newblock {\em Facial Action Coding System: A Technique for the Measurement of
  Facial Movement}.
\newblock Consulting Psychologists Press, 1978.

\bibitem{flameWEB}
models FLAME~website: dataset and code.
\newblock \url{http://flame.is.tue.mpg.de}.

\bibitem{Freifeld:ECCV:2012}
Oren Freifeld and Michael~J. Black.
\newblock Lie bodies: A manifold representation of {3D} human shape.
\newblock In {\em ECCV}, 2012.

\bibitem{gall2009motion}
Juergen Gall, Carsten Stoll, Edilson De~Aguiar, Christian Theobalt, Bodo
  Rosenhahn, and Hans-Peter Seidel.
\newblock Motion capture using joint skeleton tracking and surface estimation.
\newblock In {\em CVPR}, 2009.

\bibitem{GemanMcClure1987}
Stuart Geman and Donald~E. McClure.
\newblock Statistical methods for tomographic image reconstruction.
\newblock In {\em Proceedings of the 46th Session of the International
  Statistical Institute, Bulletin of the ISI}, volume~52, 1987.

\bibitem{hasler2009statistical}
Nils Hasler, Carsten Stoll, Martin Sunkel, Bodo Rosenhahn, and Hans-Peter
  Seidel.
\newblock A statistical model of human pose and body shape.
\newblock {\em Computer Graphics Forum}, 28(2):337--346, 2009.

\bibitem{Hasler10}
Nils Hasler, Thorsten Thorm\"{a}hlen, Bodo Rosenhahn, and Hans-Peter Seidel.
\newblock Learning skeletons for shape and pose.
\newblock In {\em Proceedings of the 2010 ACM SIGGRAPH Symposium on Interactive
  3D Graphics and Games}, I3D '10, pages 23--30, New York, NY, USA, 2010. ACM.

\bibitem{hezhangetal2015a}
Kaiming He, Xiangyu Zhang, Shaoqing Ren, and Jian Sun.
\newblock Deep residual learning for image recognition.
\newblock In {\em CVPR}, 2016.

\bibitem{Hirshberg:ECCV:2012}
David~A. Hirshberg, Matthew Loper, Eric Rachlin, and Michael~J. Black.
\newblock Coregistration: Simultaneous alignment and modeling of articulated
  {3D} shape.
\newblock In {\em ECCV}, 2012.

\bibitem{MuVS:3DV:2017}
Yinghao Huang, Federica Bogo, Christoph Lassner, Angjoo Kanazawa, Peter~V.
  Gehler, Javier Romero, Ijaz Akhter, and Michael~J. Black.
\newblock Towards accurate marker-less human shape and pose estimation over
  time.
\newblock In {\em 3DV}, 2017.

\bibitem{insafutdinov2016deepercut}
Eldar Insafutdinov, Leonid Pishchulin, Bjoern Andres, Mykhaylo Andriluka, and
  Bernt Schiele.
\newblock Deepercut: A deeper, stronger, and faster multi-person pose
  estimation model.
\newblock In {\em ECCV}, 2016.

\bibitem{ionescupapavaetal2014}
Catalin Ionescu, Dragos Papava, Vlad Olaru, and Cristian Sminchisescu.
\newblock Human3.6{M}: Large scale datasets and predictive methods for 3{D}
  human sensing in natural environments.
\newblock {\em PAMI}, 36(7):1325--1339, 2014.

\bibitem{johnson2010clustered}
Sam Johnson and Mark Everingham.
\newblock Clustered pose and nonlinear appearance models for human pose
  estimation.
\newblock In {\em BMVC}, 2010.

\bibitem{johnson2011learning}
Sam Johnson and Mark Everingham.
\newblock Learning effective human pose estimation from inaccurate annotation.
\newblock In {\em CVPR}, 2011.

\bibitem{CMU:ICCV:2015}
Hanbyul Joo, Hao Liu, Lei Tan, Lin Gui, Bart Nabbe, Iain Matthews, Takeo
  Kanade, Shohei Nobuhara, and Yaser Sheikh.
\newblock Panoptic studio: A massively multiview system for social motion
  capture.
\newblock In {\em ICCV}, 2015.

\bibitem{joo2018total}
Hanbyul Joo, Tomas Simon, and Yaser Sheikh.
\newblock Total capture: A 3{D} deformation model for tracking faces, hands,
  and bodies.
\newblock In {\em CVPR}, 2018.

\bibitem{kanazawa2018end}
Angjoo Kanazawa, Michael~J Black, David~W Jacobs, and Jitendra Malik.
\newblock End-to-end recovery of human shape and pose.
\newblock In {\em CVPR}, 2018.

\bibitem{Karras:2012:MPC}
Tero Karras.
\newblock Maximizing parallelism in the construction of {BVH}s, {O}ctrees, and
  {K}-d trees.
\newblock In {\em Proceedings of the Fourth ACM SIGGRAPH / Eurographics
  Conference on High-Performance Graphics}, pages 33--37, 2012.

\bibitem{MSR_2015_CVPR_learnShapeModel}
Sameh Khamis, Jonathan Taylor, Jamie Shotton, Cem Keskin, Shahram Izadi, and
  Andrew Fitzgibbon.
\newblock Learning an efficient model of hand shape variation from depth
  images.
\newblock In {\em CVPR}, 2015.

\bibitem{kingmawelling2013}
Diederik~P Kingma and Max Welling.
\newblock Auto-encoding variational bayes.
\newblock In {\em ICLR}, 2014.

\bibitem{lassner2017unite}
Christoph Lassner, Javier Romero, Martin Kiefel, Federica Bogo, Michael~J
  Black, and Peter~V Gehler.
\newblock Unite the people: Closing the loop between 3{D} and 2{D} human
  representations.
\newblock In {\em CVPR}, 2017.

\bibitem{LBS_PoseSpace}
John~P. Lewis, Matt Cordner, and Nickson Fong.
\newblock Pose space deformation: A unified approach to shape interpolation and
  skeleton-driven deformation.
\newblock In {\em ACM Transactions on Graphics (SIGGRAPH)}, pages 165--172,
  2000.

\bibitem{li2017learning}
Tianye Li, Timo Bolkart, Michael~J Black, Hao Li, and Javier Romero.
\newblock Learning a model of facial shape and expression from 4{D} scans.
\newblock {\em ACM Transactions on Graphics (TOG)}, 36(6):194, 2017.

\bibitem{liangxuetal2017}
Xiaodan Liang, Chunyan Xu, Xiaohui Shen, Jianchao Yang, Si Liu, Jinhui Tang,
  Liang Lin, and Shuicheng Yan.
\newblock Human parsing with contextualized convolutional neural network.
\newblock In {\em ICCV}, 2015.

\bibitem{linmaireetal2014}
Tsung-Yi Lin, Michael Maire, Serge Belongie, James Hays, Pietro Perona, Deva
  Ramanan, Piotr Doll{\'a}r, and C~Lawrence Zitnick.
\newblock Microsoft {COCO}: Common objects in context.
\newblock In {\em ECCV}, 2014.

\bibitem{Juergen_NewJournal}
Yebin Liu, Juergen Gall, Carsten Stoll, Qionghai Dai, Hans-Peter Seidel, and
  Christian Theobalt.
\newblock Markerless motion capture of multiple characters using multiview
  image segmentation.
\newblock {\em PAMI}, 35(11):2720--2735, 2013.

\bibitem{lopermahmoodetal2014}
Matthew Loper, Naureen Mahmood, and Michael~J Black.
\newblock {MoSh}: Motion and shape capture from sparse markers.
\newblock {\em ACM Transactions on Graphics (TOG)}, 33(6):220, 2014.

\bibitem{SMPL:2015}
Matthew Loper, Naureen Mahmood, Javier Romero, Gerard Pons-Moll, and Michael~J.
  Black.
\newblock {SMPL}: A skinned multi-person linear model.
\newblock {\em ACM Transactions on Graphics, (Proc. SIGGRAPH Asia)},
  34(6):248:1--248:16, Oct. 2015.

\bibitem{opendr}
Matthew~M Loper and Michael~J Black.
\newblock Open{DR}: An approximate differentiable renderer.
\newblock In {\em ECCV}, 2014.

\bibitem{amass2019}
Naureen Mahmood, Nima Ghorbani, Nikolaus~F. Troje, Gerard Pons-Moll, and
  Michael~J. Black.
\newblock {AMASS}: Archive of motion capture as surface shapes.
\newblock {\em arXiv:1904.03278}, 2019.

\bibitem{manoWEB}
MANO, models SMPL+H~website: dataset, and code.
\newblock \url{http://mano.is.tue.mpg.de}.

\bibitem{Melax:2013:handPhysics}
Stan Melax, Leonid Keselman, and Sterling Orsten.
\newblock Dynamics based {3D} skeletal hand tracking.
\newblock In {\em Graphics Interface}, pages 63--70, 2013.

\bibitem{Review_Moeslund_2006}
Thomas~B. Moeslund, Adrian Hilton, and Volker Kr\"{u}ger.
\newblock A survey of advances in vision-based human motion capture and
  analysis.
\newblock {\em CVIU}, 104(2):90--126, 2006.

\bibitem{Murray_MathInRob}
Richard~M. Murray, Li Zexiang, and S.~Shankar Sastry.
\newblock {\em A Mathematical Introduction to Robotic Manipulation}.
\newblock CRC press, 1994.

\bibitem{nocedal2006nonlinear}
Jorge Nocedal and Stephen~J Wright.
\newblock {\em Nonlinear Equations}.
\newblock Springer, 2006.

\bibitem{Lepetit:ICCV:2015:handCnnLoop}
Markus Oberweger, Paul Wohlhart, and Vincent Lepetit.
\newblock Training a feedback loop for hand pose estimation.
\newblock In {\em ICCV}, 2015.

\bibitem{OikonomidisBMVC}
Iason Oikonomidis, Nikolaos Kyriazis, and Antonis~A. Argyros.
\newblock Efficient model-based {3D} tracking of hand articulations using
  {Kinect}.
\newblock In {\em BMVC}, 2011.

\bibitem{omran2018neural}
Mohamed Omran, Christoph Lassner, Gerard Pons-Moll, Peter~V Gehler, and Bernt
  Schiele.
\newblock Neural body fitting: Unifying deep learning and model-based human
  pose and shape estimation.
\newblock In {\em 3DV}, 2018.

\bibitem{OpenPoseWEB}
OpenPose.
\newblock \url{https://github.com/CMU-Perceptual-Computing-Lab/openpose}.

\bibitem{panteleris:WACV:2018}
Paschalis Panteleris, Iason Oikonomidis, and Antonis Argyros.
\newblock Using a single {RGB} frame for real time 3{D} hand pose estimation in
  the wild.
\newblock In {\em WACV}, 2018.

\bibitem{pavlakos2018learning}
Georgios Pavlakos, Luyang Zhu, Xiaowei Zhou, and Kostas Daniilidis.
\newblock Learning to estimate 3{D} human pose and shape from a single color
  image.
\newblock In {\em CVPR}, 2018.

\bibitem{BFM2009}
Pascal Paysan, Reinhard Knothe, Brian Amberg, Sami Romdhani, and Thomas Vetter.
\newblock A 3{D} face model for pose and illumination invariant face
  recognition.
\newblock In {\em 2009 Sixth IEEE International Conference on Advanced Video
  and Signal Based Surveillance}, pages 296--301, 2009.

\bibitem{Dyna:SIGGRAPH:2015}
Gerard Pons-Moll, Javier Romero, Naureen Mahmood, and Michael~J. Black.
\newblock Dyna: A model of dynamic human shape in motion.
\newblock {\em ACM Transactions on Graphics, (Proc. SIGGRAPH)},
  34(4):120:1--120:14, July 2015.

\bibitem{PonsModelBased}
Gerard Pons-Moll and Bodo Rosenhahn.
\newblock {\em Model-Based Pose Estimation}, chapter~9, pages 139--170.
\newblock Springer, 2011.

\bibitem{rhodin2018learning}
Helge Rhodin, J{\"o}rg Sp{\"o}rri, Isinsu Katircioglu, Victor Constantin,
  Fr{\'e}d{\'e}ric Meyer, Erich M{\"u}ller, Mathieu Salzmann, and Pascal Fua.
\newblock Learning monocular 3{D} human pose estimation from multi-view images.
\newblock In {\em CVPR}, 2018.

\bibitem{CAESAR}
Kathleen~M. Robinette, Sherri Blackwell, Hein Daanen, Mark Boehmer, Scott
  Fleming, Tina Brill, David Hoeferlin, and Dennis Burnsides.
\newblock {Civilian American and European Surface Anthropometry Resource
  (CAESAR)} final report.
\newblock Technical Report AFRL-HE-WP-TR-2002-0169, {US Air Force Research
  Laboratory}, 2002.

\bibitem{romero2017embodied}
Javier Romero, Dimitrios Tzionas, and Michael~J Black.
\newblock Embodied hands: Modeling and capturing hands and bodies together.
\newblock {\em ACM Transactions on Graphics (TOG)}, 2017.

\bibitem{DART-Schmidt-RSS-14}
Tanner Schmidt, Richard Newcombe, and Dieter Fox.
\newblock {DART}: Dense articulated real-time tracking.
\newblock In {\em RSS}, 2014.

\bibitem{simon2017hand}
Tomas Simon, Hanbyul Joo, Iain Matthews, and Yaser Sheikh.
\newblock Hand keypoint detection in single images using multiview
  bootstrapping.
\newblock In {\em CVPR}, 2017.

\bibitem{srinath_iccv2013}
Srinath Sridhar, Antti Oulasvirta, and Christian Theobalt.
\newblock Interactive markerless articulated hand motion tracking using {RGB}
  and depth data.
\newblock In {\em ICCV}, 2013.

\bibitem{hilton2007surface}
Jonathan Starck and Adrian Hilton.
\newblock Surface capture for performance-based animation.
\newblock {\em IEEE computer graphics and applications}, 27(3), 2007.

\bibitem{collisionDeformableObjects}
Matthias Teschner, Stefan Kimmerle, Bruno Heidelberger, Gabriel Zachmann, Laks
  Raghupathi, Arnulph Fuhrmann, Marie-Paule Cani, Fran{\c{c}}ois Faure, Nadia
  Magnenat-Thalmann, Wolfgang Strasser, and Pascal Volino.
\newblock Collision detection for deformable objects.
\newblock In {\em Eurographics}, 2004.

\bibitem{Tkach:SIGGRAPH:2016}
Anastasia Tkach, Mark Pauly, and Andrea Tagliasacchi.
\newblock Sphere-meshes for real-time hand modeling and tracking.
\newblock {\em ACM Transactions on Graphics (TOG)}, 35(6), 2016.

\bibitem{Tzionas:IJCV:2016}
Dimitrios Tzionas, Luca Ballan, Abhilash Srikantha, Pablo Aponte, Marc
  Pollefeys, and Juergen Gall.
\newblock Capturing hands in action using discriminative salient points and
  physics simulation.
\newblock {\em IJCV}, 118(2):172--193, 2016.

\bibitem{Vlasic2005}
Daniel Vlasic, Matthew Brand, Hanspeter Pfister, and Jovan Popovi{\'c}.
\newblock Face transfer with multilinear models.
\newblock {\em ACM transactions on graphics (TOG)}, 24(3):426--433, 2005.

\bibitem{wei2016convolutional}
Shih-En Wei, Varun Ramakrishna, Takeo Kanade, and Yaser Sheikh.
\newblock Convolutional pose machines.
\newblock In {\em CVPR}, 2016.

\bibitem{xu2018monoperfcap}
Weipeng Xu, Avishek Chatterjee, Michael Zollh{\"o}fer, Helge Rhodin, Dushyant
  Mehta, Hans-Peter Seidel, and Christian Theobalt.
\newblock Monoperfcap: Human performance capture from monocular video.
\newblock {\em ACM Transactions on Graphics (TOG)}, 37(2):27, 2018.

\bibitem{Yang2011}
Fei Yang, Jue Wang, Eli Shechtman, Lubomir Bourdev, and Dimitri Metaxas.
\newblock Expression flow for 3{D}-aware face component transfer.
\newblock {\em ACM Transactions on Graphics (TOG)}, 30(4):60, 2011.

\bibitem{hands:Survey:2017}
Shanxin Yuan, Guillermo Garcia-Hernando, Bj{\"o}rn Stenger, Gyeongsik Moon, Ju
  Yong~Chang, Kyoung Mu~Lee, Pavlo Molchanov, Jan Kautz, Sina Honari, Liuhao
  Ge, et~al.
\newblock Depth-based 3{D} hand pose estimation: From current achievements to
  future goals.
\newblock In {\em CVPR}, 2018.

\bibitem{zollhofer2018stateFaces}
Michael Zollh{\"o}fer, Justus Thies, Pablo Garrido, Derek Bradley, Thabo
  Beeler, Patrick P{\'e}rez, Marc Stamminger, Matthias Nie{\ss}ner, and
  Christian Theobalt.
\newblock State of the art on monocular 3{D} face reconstruction, tracking, and
  applications.
\newblock {\em Computer Graphics Forum}, 37(2):523--550, 2018.

\end{thebibliography}
}

\clearpage


\section*{Supplementary material}

# Expressive Body Capture: 3D Hands, Face, and Body from a Single Image **Supplementary Material**

Georgios Pavlakos[*1,2], Vasileios Choutas[*1], Nima Ghorbani[1], Timo Bolkart[1], Ahmed A. A. Osman[1],
Dimitrios Tzionas[1], and Michael J. Black[1]

[1]MPI for Intelligent Systems, Tübingen, DE , [2] University of Pennsylvania, PA, USA
{gpavlakos, vchoutas, nghorbani, tbolkart, aosman, dtzionas, black}@tuebingen.mpg.de

## 1. Qualitative results

**Comparison of SMPL, SMPL+H & SMPL-X:** In Section 4.2 of the main paper, in Table 1 we present a quantitative comparison between different models with different modeling capacities. In Fig. A.1 we present a similar comparison for SMPL (left), SMPL+H (middle) and SMPL-X (right) for an image of the EHB dataset. For fair comparison we fit all models with a variation of SMPLify-X to a single RGB image. The figure reflects the same findings as Table 1 of the paper, but qualitatively; there is a clear increase in expressiveness from left to right, as model gets richer from body-only (SMPL) to include hands (SMPL+H) or hands and face (SMPL-X).

**Holistic vs part models:** In Section 4.2 and Fig. 5 of the main paper we compare our holistic SMPL-X model to the hand-only approach of [24] on EHB. Figure A.2 shows a similar qualitative comparison, this time on the data of [24]. To further explore the benefit of holistic reasoning, we also focus on the head and we compare SMPL-X fitting to a head-only method by fitting FLAME [16] to 2D keypoints similar to our method. The context of the full body stabilizes head estimation for occlusions or non-frontal views (see Fig. A.3). This benefit is also quantitative, where the holistic SMPL-X improves over the head-only fitting by 17% in our EHF dataset in terms of vertex-to-vertex error.

**Failure cases:** Figure A.4 shows some representative failure cases; depth ambiguities can cause wrong estimation of torso pose or wrong ordinal depth estimation of body parts due to the simple 2D re-projection data term. Furthermore, occluded joints leave certain body parts unconstrained, which currently leads to failures. We plan to address this in future work, by employing a visibility term in the objective.

## 2. Collision Penalizer

In Section 3.4 of the paper we describe the collision penalizer. For technical details and visualizations the reader

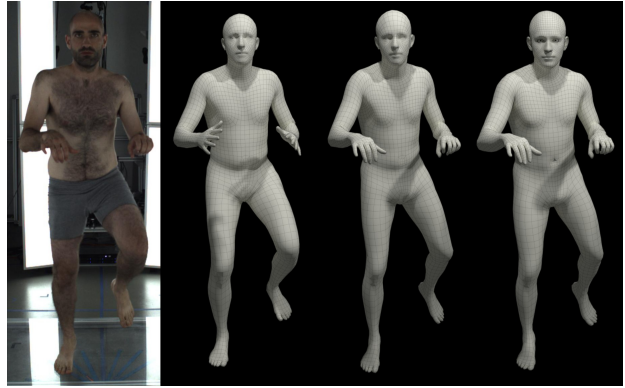

Figure A.1. Comparison of SMPL (left), SMPL+H (middle) and SMPL-X (right) on the EHB dataset, using the male models. For fair comparison we fit all models with a variation of SMPLify-X to a single RGB image. The results show a clear increase in *expressiveness* from let to right, as model gets richer from body-only (SMPL) to include hands (SMPL+H) or hands and face (SMPL-X).

is redirected to [4, 28], but for the sake of completion we include the mathematical formulation also here.

We first detect a list of colliding triangles $\mathcal{C}$ by employing Bounding Volume Hierarchies (BVH) [27] and compute local conic 3D distance fields $\Psi : \mathbb{R}^3 \to \mathbb{R}_+$ defined by the triangles $\mathcal{C}$ and their normals $n \in \mathbb{R}^3$. Penetrations are then penalized by the depth of intrusion, efficiently computed by the position in the distance field. For two colliding triangles $f_s$ and $f_t$ intrusion is bi-directional; the vertices $v_t \in \mathbb{R}^3$ of $f_t$ are the *intruders* in the distance field $\Psi_{f_s}$ of the *receiver* triangle $f_s$ and are penalized by $\Psi_{f_s}(v_t)$, and vice-versa. Thus, the collision term $E_\mathcal{C}$ is defined as

$$E_\mathcal{C}(\theta) = \sum_{(f_s(\theta), f_t(\theta)) \in \mathcal{C}} \left\{ \sum_{v_s \in f_s} \| - \Psi_{f_t}(v_s) n_s \|^2 + \sum_{v_t \in f_t} \| - \Psi_{f_s}(v_t) n_t \|^2 \right\}. \quad (1)$$

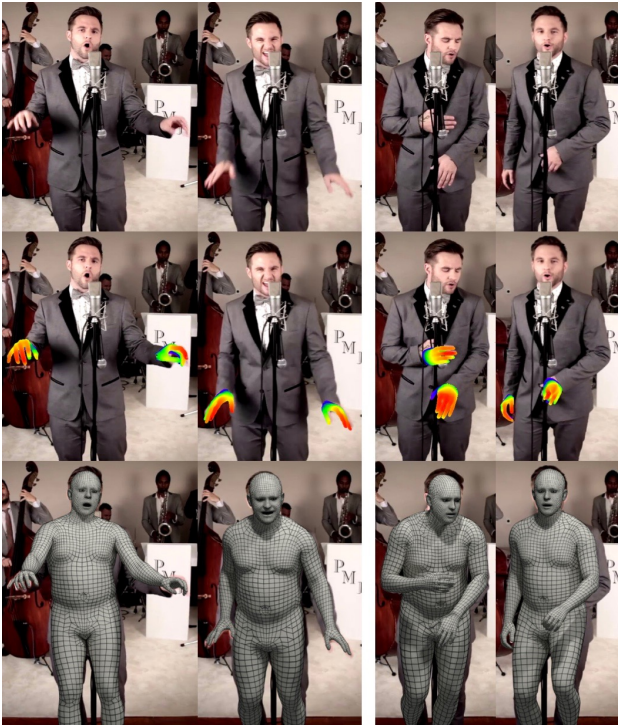

Figure A.2. Comparison of the hands-only approach of [24] (middle row) against SMPLify-X with the male SMPL-X (bottom row). Both approaches depend on OpenPose [23]. In case of good 2D detections both perform well (left group). In case of noisy detections (right group) fitting a holistic model is more robust.

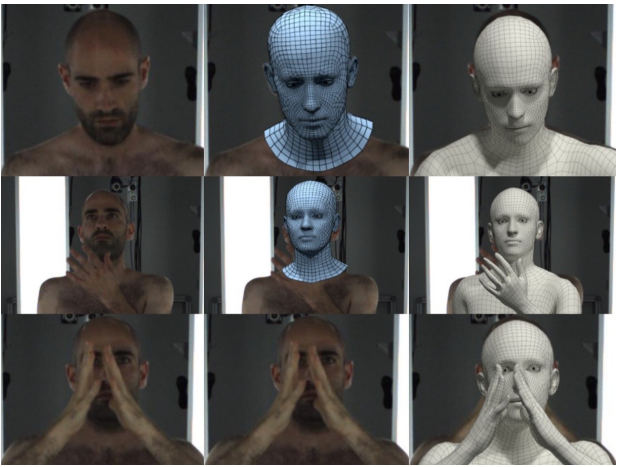

Figure A.3. Fitting SMPL-X (right) versus FLAME (middle). For minimal occlusions and frontal views (top) both methods perform well. For moderate (middle) or extreme (bottom) occlusions the body provides crucial context and improves fitting (bottom: missing FLAME model indicates a complete fitting failure).

For the case where $f_t$ is the *intruder* and $f_s$ is the *receiver* (similarly for the opposite case) the cone for the distance

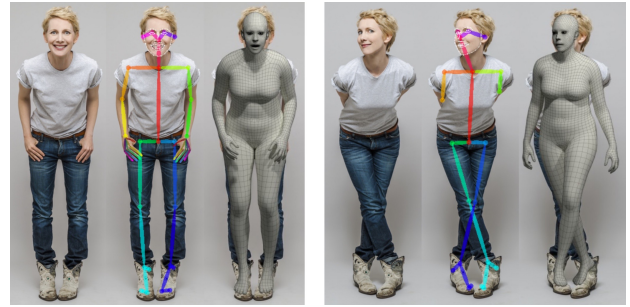

Figure A.4. Failure cases for *SMPLify-X* with the female SMPL-X for expressive RGB images similar to the ones of Figures 1 and 2 of the main paper. In the left case, 2D keypoints are reasonable, but due to depth ambiguities the torso pose is wrong, while the head shape is under-estimated. In the right case, the arms and hands are occluded and due to lack of constraints the arm and hand pose is wrong. The ordinal depth for feet is estimated wrongly, while similarly to the left case the torso pose and head shape are not estimated correctly. *Left*: Input RGB image. *Middle*: Intermediate 2D keypoints from OpenPose. *Right*: SMPL-X fittings overlaid on the RGB image.

field $\Psi_{f_s}$ is defined as

$$\Psi_{f_s}(v_t) = \begin{cases} |(1-\Phi(v_t))\Upsilon(n_{f_s}\cdot(v_t-\mathbf{o}_{f_s}))|^2 & \Phi(v_t) < 1 \\ 0 & \Phi(v_t) \geq 1 \end{cases} \quad (2)$$

where $\mathbf{o}_{f_s} \in \mathbb{R}^3$ is the circumcenter and $r_{f_s} \in \mathbb{R}_{>0}$ the radius of the circumcircle for the *receiver* triangle. The term

$$\Phi(v_t) = \frac{\|(v_t - \mathbf{o}_{f_s}) - (n_{f_s}\cdot(v_t - \mathbf{o}_{f_s}))n_{f_s}\|}{-\frac{r_{f_s}}{\sigma}(n_{f_s}\cdot(v_t - \mathbf{o}_{f_s})) + r_{f_s}} \quad (3)$$

projects the vertex $v_t$ onto the axis of the cone defined by the triangle normal $n_{f_s}$ and going through the circumcenter $\mathbf{o}_{f_s}$. It then measures the distance to it, scaled by the radius of the cone at this point. If $\Phi(v) < 1$ the vertex is inside the cone and if $\Phi(v) = 0$ the vertex is on the axis. The term

$$\Upsilon(x) = \begin{cases} -x + 1 - \sigma & x \leq -\sigma \\ -\frac{1-2\sigma}{4\sigma^2}x^2 - \frac{1}{2\sigma}x + \frac{1}{4}(3-2\sigma) & x \in (-\sigma, +\sigma) \\ 0 & x \geq +\sigma \end{cases} \quad (4)$$

measures how far the projected point is from the circumcenter to define the intensity of penalization. For $\Upsilon(x) < 0$ the projected point is behind the triangle. For $x \in (-\sigma, +\sigma)$ the penalizer is quadratic, while for $x > |\sigma|$ it becomes linear. The parameter $\sigma$ also defines the field of view of the cone. In contrast to [4, 28] that use $mm$ unit and $\sigma = 0.5$, we use $m$ unit and $\sigma = 0.0001$. For the resolution of our meshes, we empirically find that this value allows for both penalizing penetrations, as well as for not over-penalizing in case of self-contact, *e.g.* arm resting on knee.

As seen in Fig. A.5, for certain parts of the body, like the eyes, toes, armpits and crotch, as well as neighboring parts

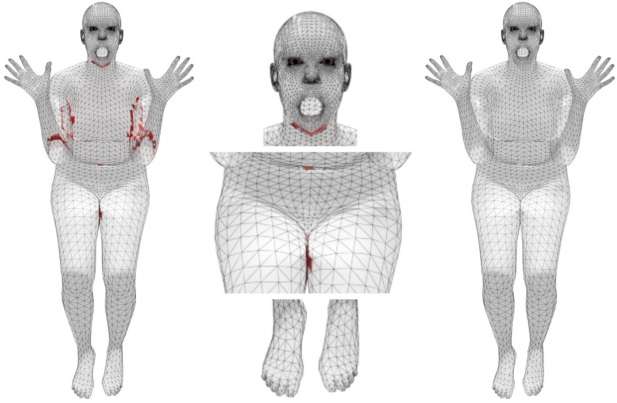

Figure A.5. For certain parts of the body, like the eyes, toes, armpits and crotch, as well as neighboring parts in the kinematic chain, there is either always or frequently self-contact. The triangles for which collisions are detected are highlighted with red (left, middle). Since the model does not model deformations due to contact, for simplicity we just ignore collisions for these areas (right).

in the kinematic chain, there is either always or frequently self-contact. For simplicity, since the model does not model deformations due to contact, we simply ignore collisions for neighboring parts in these areas. Our empirical observations suggest that collision detection for the other parts resolves most penetrations and helps prevent physically implausible poses. Figure A.6 shows the effect of the collision penalizer, by including or excluding it from optimization, and depicts representative success and failure cases.

For computational efficiency, we developed a custom PyTorch wrapper operator for our CUDA kernel based on the highly parallelized implementation of BVH [14].

## 3. Optimization

In Section 3.6 of the paper we present the main information about optimizing our objective function, while in the following we present omitted details.

To keep optimization tractable, we use a PyTorch implementation and the Limited-memory BFGS optimizer (L-BFGS) [22] with strong Wolfe line search. We use a learning rate of $1.0$ and $30$ maximum iterations. For the annealing scheme presented in Section 3.6 we take the following three steps. We start with high regularization to mainly refine the global body pose, $(\gamma_b = 1, \gamma_h = 0, \gamma_f = 0)$ and gradually increase the influence of hand keypoints to refine the pose of the arms $(\gamma_b = 1, \gamma_h = 0.1, \gamma_f = 0)$. After converging to a better pose estimate, we increase the influence of both hands and facial keypoints to capture expressivity $(\gamma_b = 1, \gamma_h = 2, \gamma_f = 2)$. Throughout the above steps the weights $\lambda_\alpha, \lambda_\beta, \lambda_\mathcal{E}$ in the objective function $E$ start with high regularization that progressively lowers to allow for better fitting. The only exception is $\lambda_\mathcal{C}$ that progressively

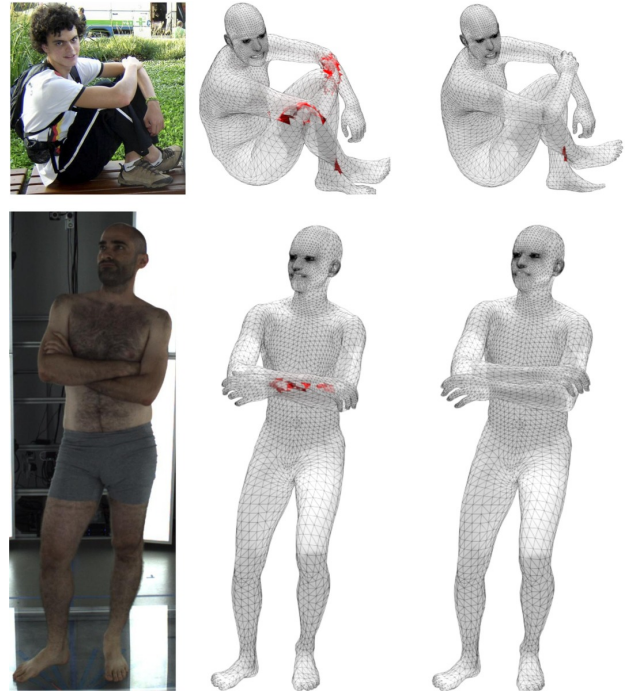

Figure A.6. Effect of the collision penalizer. The colliding triangles are highlighted to show penetrations at the end of optimization with SMPLify-X without (middle) and with (right) the collision term in the objective function. The top row shows a successful case, were optimization resolves most collisions and converges in a physically plausible pose that reflects the input image. The bottom row shows a failure case, for which arm crossing causes a lot of collisions due to self-touch. The final pose (right) is still physically plausible, but optimization gets trapped in a local minima and the pose does not reflect the input image.

increases while the influence of hands and facial keypoints gets stronger in $E_J$, thus bigger pose changes and more collisions are expected.

Regarding the weights of the optimization, they are set empirically and the exact parameters for each stage of the optimization will be released with our code. For more intuition we performed sensitivity analysis by perturbing each weight $\lambda$ separately by up to $\pm 25\%$. This resulted to relative changes smaller than $6\%$ in the vertex-to-vertex error metric, meaning that our approach is robust for significant weight ranges and not sensitive to fine-tuning. The detailed results are presented in Fig. A.7.

## 4. Quantitative evaluation on "Total Capture"

In the main paper we present a curated dataset called *Expressive hands and faces dataset (EHF)* with ground-truth shape for bodies, hands and faces together.

Since the most relevant model is Frank [13], we also use the "Total Capture" dataset [8] of the authors, focusing on the "PtCloudDB" part that includes pseudo ground-truth for

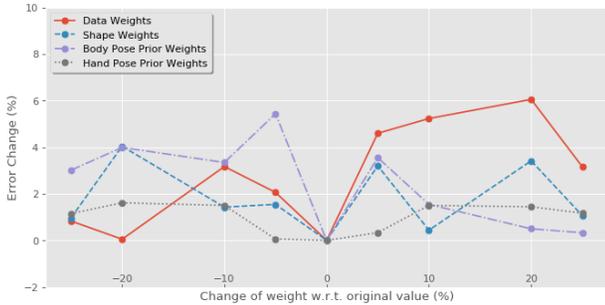

Figure A.7. Sensitivity of the weights for the different terms of the optimization. Each weight $\lambda$ is perturbed separately up to $\pm 25\%$. The relative changes in the vertex-to-vertex error are smaller than $6\%$, indicating that our approach is robust for significant weight ranges and not sensitive to fine-tuning.

|  |  | SMPLify-X using | |
| --- | --- | --- | --- |
| Error Joints | Alignment Joints | GT 2D | pred 2D |
| Body | Body | 92.6 | 117.5 |
| Body+H+F | Body | 101.2 | 136.2 |
| Body+H+F | Body+H+F | 71.2 | 93.4 |

Table A.1. Quantitative results on the selected frames from CMU Panoptic Studio, using SMPLify-X on the 2D re-projection of the ground-truth 3D joints, and the 2D joints detected by OpenPose respectively. The numbers are mean 3D joint errors after Procrustes alignment. First, we evaluate the error on the body-only keypoints after Procrustes alignment with the ground-truth body-only keypoints (row 1). Then, we consider the same alignment using body-only keypoints, but we evaluate the joint error across all the body+hands+face keypoints (row 2). Finally, we align the prediction using all body+hands+face keypoints and we report the mean error across all of them (row 3).

all body, face and hands. This pseudo ground-truth is created with triangulated 3D joint detection from multi-view with OpenPose [23]. We curate and pick 200 images, according to the degree of visibility of the body in the image, interesting hand poses and facial expressions. In the following, we refer to this data as *"total hands and faces"* *(THF)* dataset. Figure A.8 shows qualitative results on part of THF. For each group of images the top row shows a reference RGB image, the middle row shows SMPLify-X results using pseudo ground-truth OpenPose keypoints (projected on 2D for use by our method), while the bottom row shows SMPLify-X results using 2D OpenPose keypoints estimated with [23]. Quantitative results for this dataset are reported in Table A.1.

## 5. Quantitative evaluation on Human3.6M

In the main manuscript (Table 1), we demonstrated that evaluating the reconstruction accuracy using 3D body joints is not representative of the accuracy and the detail of a

| Method | Mean (mm) | Median (mm) |
| --- | --- | --- |
| SMPLify [5] | 82.3 | 69.3 |
| SMPLify-X | 75.9 | 60.8 |

Table A.2. Quantitative results on the Human3.6M dataset [10]. The numbers are mean 3D joint errors after Procrustes alignment. We use the evaluation protocol of [5].

method's reconstruction. However, many approaches do evaluate quantitatively based on 3D body joints metrics, so here we compare our results with SMPLify [5] to demonstrate that our approach is not only more natural, expressive and detailed, but the results are also more accurate in the common metrics. In Table A.2 we present our results using the Human3.6M [10] dataset. We follow the same protocol as [5] and we report results after Procrustes alignment with the ground-truth 3D pose. Even though there are several factors that improve our approach over SMPLify and this experiment does not say which is more important (we direct the reader to the ablative study in Table 2 of the main manuscript for this), we still outperform the original SMPLify using this crude metric based on 3D joints.

## 6. Qualitative evaluation on MPII

In Fig. A.14 we present qualitative results on the MPII dataset [3]. For this dataset we also include some cases with low resolution, heavily occluded or cropped people.

## 7. Model

In Section 3.1 of the main manuscript we describe the SMPL-X model. The model shape space is trained on the CAESAR database [26]. In Fig. A.9 we present the percentage of explained variance as a function of the number of PCA components used. All models explain more than $95\%$ of the variance with 10 principle components.

We further evaluate the model on a held out set of 180 alignments of male and female subjects in different poses. The male model is evaluated on the male alignments, the female model is evaluated on the female alignments, while the gender neutral is evaluated on both male and female alignments. We report the model alignment vertex-to-vertex (v2v) mean absolute error as a function of the number of principle components used, shown in Fig. A.10.

## 8. VPoser

In Section 3.3 of the main manuscript we introduce a new parametrization of the human pose and a prior on this parameterization, also referred to as VPoser. In this Section we present further details on the data preparation and implementation.

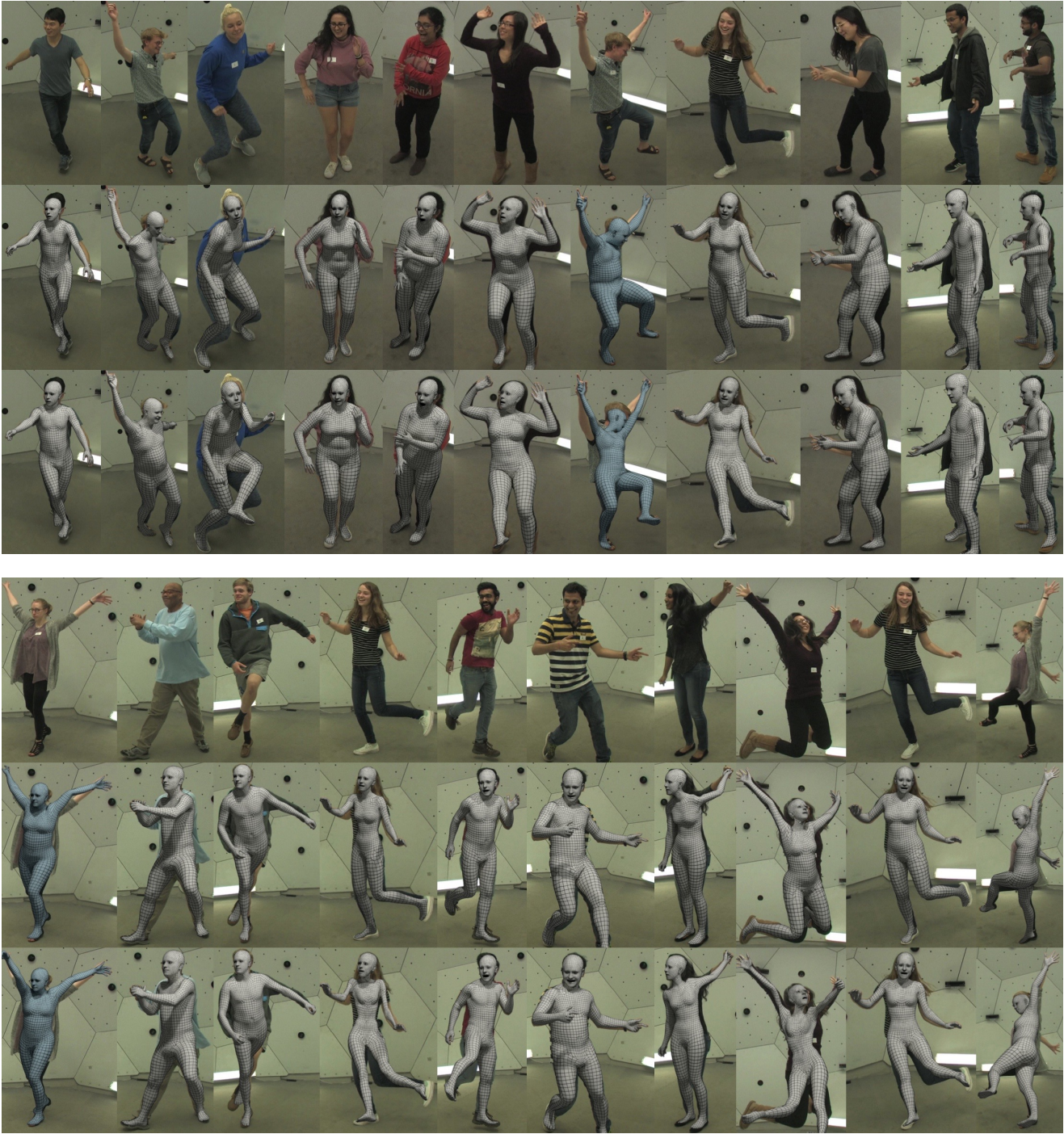

Figure A.8. Qualitative results on some of the data of the "total capture" dataset [8], focusing on the "PtCloudDB" part that includes pseudo ground-truth for all body, face and hands. We curate and pick 200 images, according to degree of body coverage in the image and interesting hand poses and facial expressions. We refer to this data as "total hands and faces" dataset (THF). *Top row*: Reference RGB image. *Middle row*: SMPLify-X results using pseudo ground-truth OpenPose keypoints (3D keypoints of [8] estimated from multi-view and projected on 2D). *Bottom row*: SMPLify-X results using 2D OpenPose keypoints estimated with [23]. *Gray* color depicts the gender-specific model for confident gender detections. *Blue* is the gender-neutral model that is used when the gender classifier is uncertain.

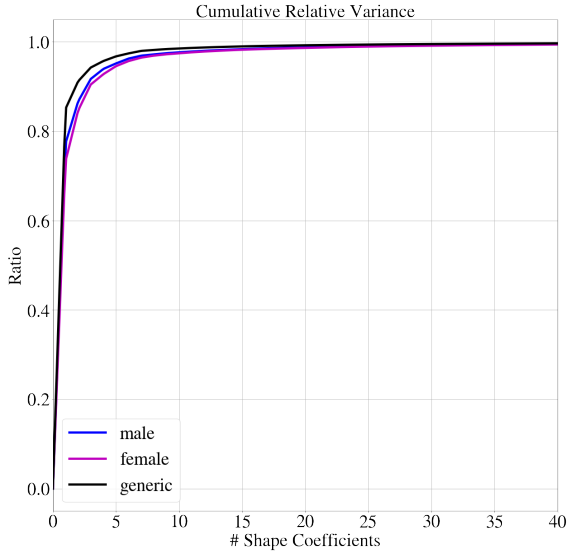

Figure A.9. Cumulative relative variance of the CAESAR dataset explained as a function of the number of shape coefficients for three SMPL-X models: male, female, gender neutral model.

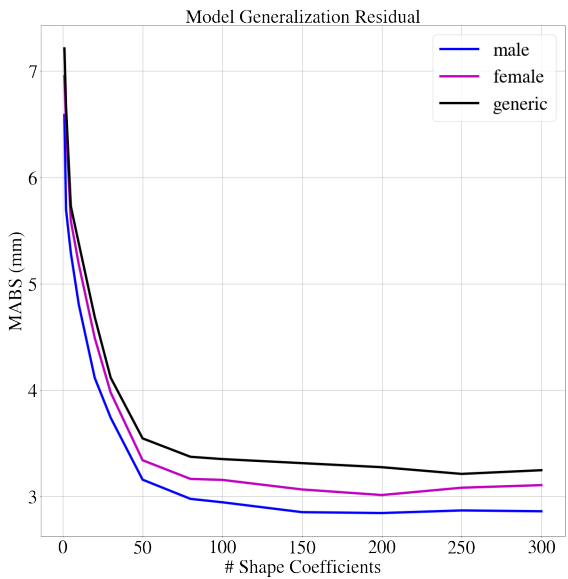

Figure A.10. Evaluating SMPL-X generalization on a held out test set of male and female 3D alignments.

### 8.1. Data preparation

We use SMPL body pose parameters extracted with [19, 21] from human motion sequences of CMU [7], Human3.6M [10], and PosePrior [2] as our dataset. Subsequently, we hold out parameters for Subjects 9 and 11 of Human3.6M as our test set. We randomly select $5\%$ of the training set as our validation set and use that to make snapshots of the model with minimum validation loss. We choose matrix rotations for our pose parameterization.

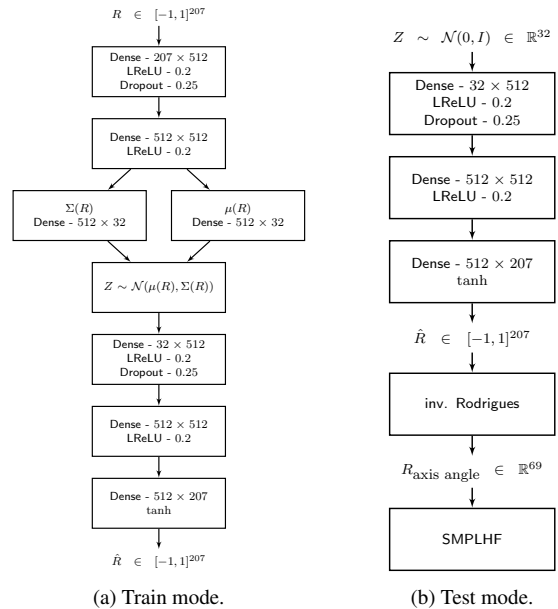

Figure A.11. VPoser model in different modes. For training the network consists of an encoder and a decoder. For testing we use the latent code instead of the body pose parameters, *i.e.* $\theta_b$, of SMPL-X, which are described in Section 3.1 of the main paper. By "inverse Rodrigues" we note the conversion from a rotation matrix to an axis-angle representation for posing SMPL-X.

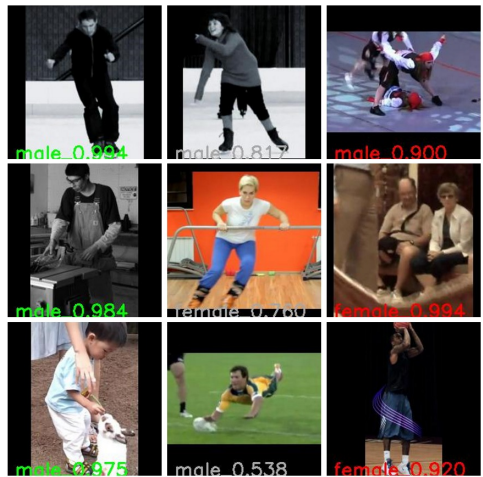

Figure A.12. Gender classifier results on the test set. From left to right column: Successful predictions, predictions discarded due to low confidence($< 0.9$), failure cases.

### 8.2. Implementation details

For implementation we use TensorFlow [1] and later port the trained model and weights to PyTorch [25]. Figure A.11 shows the network architecture during training and test time. We use only fully-connected layers, with LReLU [20] non-linearity and keep the encoder and decoder symmet-

ric. The encoder has two dense layers with 512 units each, and then one dense layer for mean and another for variance of the VAE's posterior Normal distribution. The decoder weights have the same shape as the encoder, only in reverse order. We use the ADAM solver [15], and update the weights of the network to minimize the loss defined in Eq. 5 of the main manuscript. We empirically choose the values for loss weights as: $c_1 = 0.005, c_2 = 1.0 - c_2, c_3 = 1.0, c_4 = 1.0, c_5 = 0.0005$. We train for 60 epochs for each of the following learning rates: $[5e-4, 1e-4, 5e-5]$.

After training, the latent space describes a manifold of physically plausible human body poses, that can be used for efficient 2D-to-3D lifting. Figure A.13 shows a number of random samples drawn from the latent space of the model.

## 9. Gender lassifier

Figure A.12 shows some qualitative results of the gender classifier on the test set.

### 9.1. Training data

For training data we employ the LSP [11], LSP-extended [12], MPII [3], MS-COCO [18], LIP [17] datasets, respecting their original train and test splits. To curate our data for gender annotations we collect tight crops around persons and keep only the ones for which there is at least one visible joint with high confidence for the head, torso and for each limb. We further reject crops with size smaller than $200 \times 200$ pixels. The gathered samples are annotated with gender labels using Amazon Mechanical Turk. Each image is annotated by two Turkers and we keep only the ones with consistent labels.

### 9.2. Implementation details

For implementation we use Keras [6] with TensorFlow [1] backend. We use a pretrained ResNet18 [9] for feature extraction and append fully-connected layers for our classifier. We employ a cross entropy loss, augmented with an L2 norm on the weights. Each data sample is resized to $224 \times 224$ pixels to be compatible with the ResNet18 [9] architecture. We start by training the final fully-connected layers for two epochs with each of the following learning rate values $[1e-3, 1e-4, 1e-5, 1e-6]$. Afterwards, the entire network is finetuned end-to-end for two epochs using these learning rates $[5e-5, 1e-5, 1e-6, 1e-7]$. Optimization is performed using Adam [15].

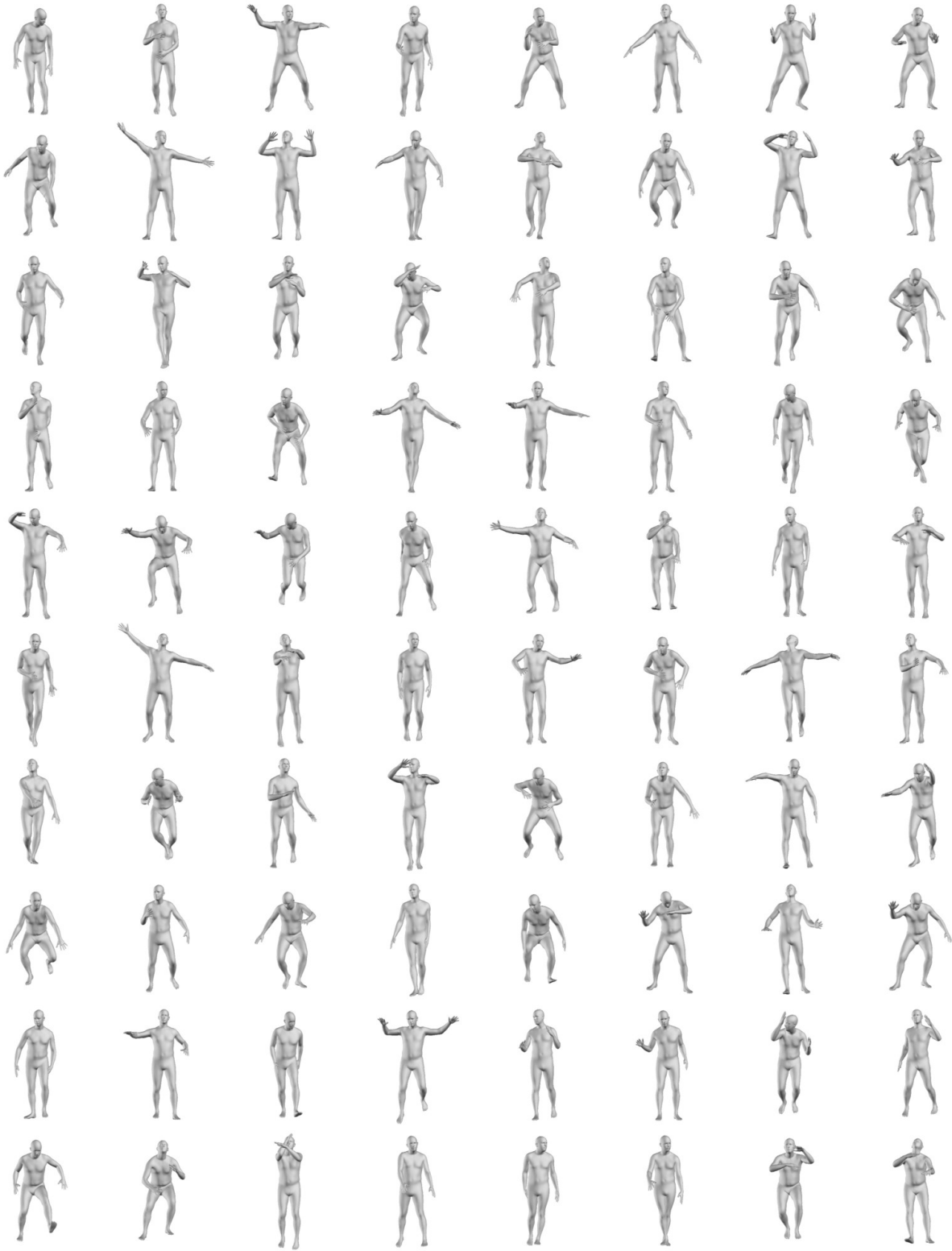

Figure A.13. Random pose samples from the latent space of VPoser. We sample from a 32 dimensional normal distribution and feed the value to the decoder of VPoser; shown in Figure A.11b. SMPL is then posed with the decoder output, after conversion to an axis-angle representation.

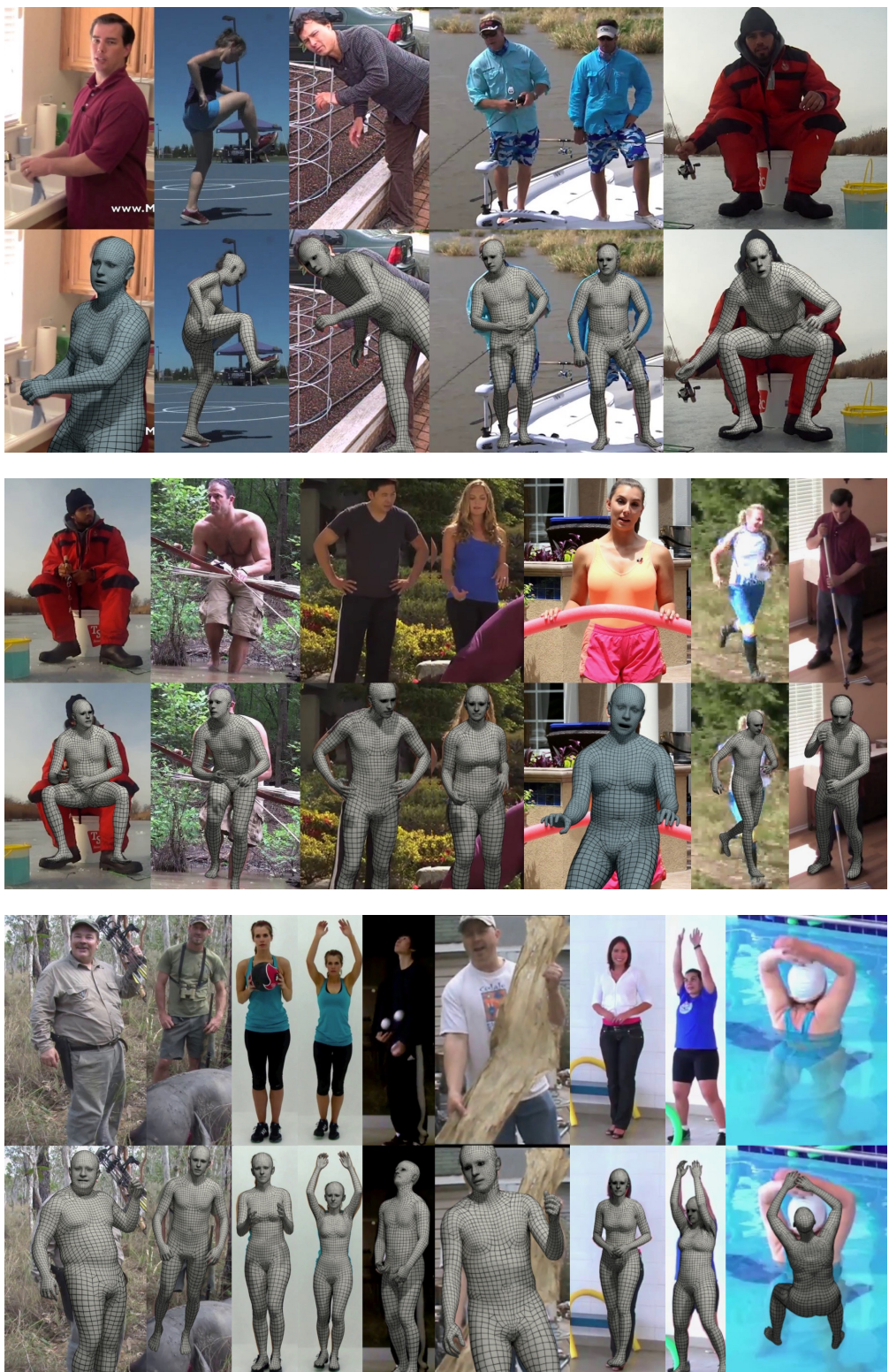

Figure A.14. Qualitative results of SMPLify-X with SMPL-X on the MPII dataset [3]. In this figure we also include images with some heavily occluded or cropped bodies. *Gray* color depicts the gender-specific model for confident gender detections. *Blue* is the gender-neutral model that is used when the gender classifier is uncertain or when cropping does not agree with the filtering criterion described in ubsection 9.1.

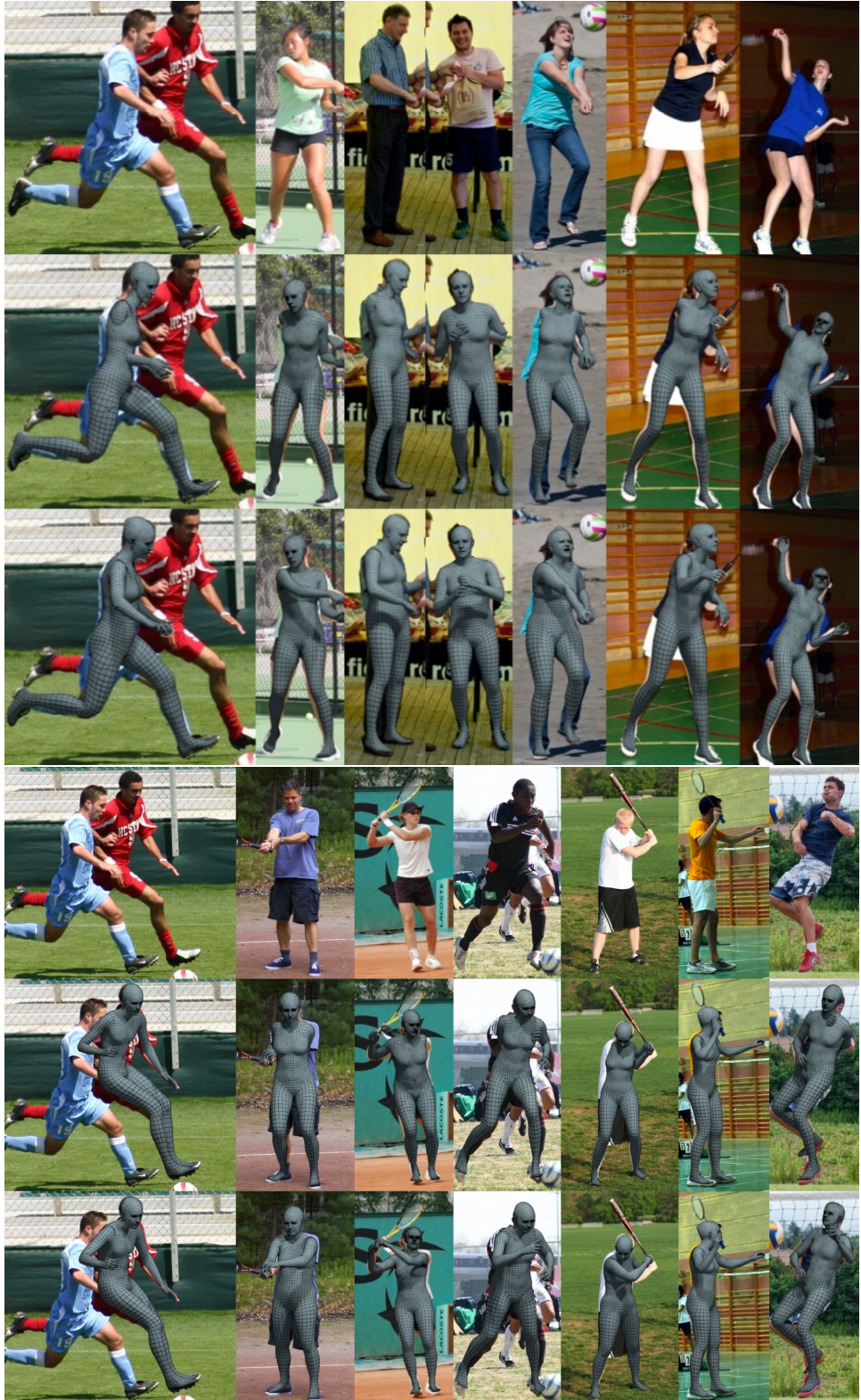

Figure A.15. Results of SMPLify-X fitting for the LSP dataset. For each group of images we compare two body priors; the top row shows a reference RGB image, the bottom row shows results of SMPLify with VPoser, while the middle row shows results for which VPoser is replaced with the GMM body pose prior of SMPLify [5]. To eliminate factors of variation, for this comparison we use the gender neutral SMPL-X model.

## References

[1] Martín Abadi, Paul Barham, Jianmin Chen, Zhifeng Chen, Andy Davis, Jeffrey Dean, Matthieu Devin, Sanjay Ghemawat, Geoffrey Irving, Michael Isard, et al. Tensorflow: A system for large-scale machine learning. In *OSDI*, volume 16, pages 265–283, 2016. 6, 7

[2] Ijaz Akhter and Michael J. Black. Pose-conditioned joint angle limits for 3D human pose reconstruction. In *CVPR*, 2015. 6

[3] Mykhaylo Andriluka, Leonid Pishchulin, Peter Gehler, and Bernt Schiele. 2D human pose estimation: New benchmark and state of the art analysis. In *CVPR*, 2014. 4, 7, 9

[4] Luca Ballan, Aparna Taneja, Juergen Gall, Luc Van Gool, and Marc Pollefeys. Motion capture of hands in action using discriminative salient points. In *ECCV*, 2012. 1, 2

[5] Federica Bogo, Angjoo Kanazawa, Christoph Lassner, Peter Gehler, Javier Romero, and Michael J Black. Keep it SMPL: Automatic estimation of 3D human pose and shape from a single image. In *ECCV*, 2016. 4, 10

[6] François Chollet et al. Keras. https://keras.io, 2015. 7

[7] CMU. CMU MoCap dataset. 6

[8] Total Capture Dataset. http://domedb.perception.cs.cmu.edu. 3, 5

[9] Kaiming He, Xiangyu Zhang, Shaoqing Ren, and Jian Sun. Deep residual learning for image recognition. In *CVPR*, 2016. 7

[10] Catalin Ionescu, Dragos Papava, Vlad Olaru, and Cristian Sminchisescu. Human3.6M: Large scale datasets and predictive methods for 3D human sensing in natural environments. *PAMI*, 36(7):1325–1339, 2014. 4, 6

[11] Sam Johnson and Mark Everingham. Clustered pose and nonlinear appearance models for human pose estimation. In *BMVC*, 2010. 7

[12] Sam Johnson and Mark Everingham. Learning effective human pose estimation from inaccurate annotation. In *CVPR*, 2011. 7

[13] Hanbyul Joo, Tomas Simon, and Yaser Sheikh. Total capture: A 3D deformation model for tracking faces, hands, and bodies. In *CVPR*, 2018. 3

[14] Tero Karras. Maximizing parallelism in the construction of BVHs, Octrees, and K-d trees. In *Proceedings of the Fourth ACM SIGGRAPH / Eurographics Conference on High-Performance Graphics*, pages 33–37, 2012. 3

[15] Diederik P Kingma and Jimmy Ba. Adam: A method for stochastic optimization. In *ICLR*, 2015. 7

[16] Tianye Li, Timo Bolkart, Michael J Black, Hao Li, and Javier Romero. Learning a model of facial shape and expression from 4D scans. *ACM Transactions on Graphics (TOG)*, 36(6):194, 2017. 1

[17] Xiaodan Liang, Chunyan Xu, Xiaohui Shen, Jianchao Yang, Si Liu, Jinhui Tang, Liang Lin, and Shuicheng Yan. Human parsing with contextualized convolutional neural network. In *ICCV*, 2015. 7

[18] Tsung-Yi Lin, Michael Maire, Serge Belongie, James Hays, Pietro Perona, Deva Ramanan, Piotr Dollár, and C Lawrence Zitnick. Microsoft COCO: Common objects in context. In *ECCV*, 2014. 7

[19] Matthew Loper, Naureen Mahmood, and Michael J Black. MoSh: Motion and shape capture from sparse markers. *ACM Transactions on Graphics (TOG)*, 33(6):220, 2014. 6

[20] Andrew L Maas, Awni Y Hannun, and Andrew Y Ng. Rectifier nonlinearities improve neural network acoustic models. In *ICML Workshops*, 2013. 6

[21] Naureen Mahmood, Nima Ghorbani, Nikolaus F. Troje, Gerard Pons-Moll, and Michael J. Black. AMASS: Archive of motion capture as surface shapes. *arXiv:1904.03278*, 2019. 6

[22] Jorge Nocedal and Stephen Wright. *Numerical Optimization*. Springer, New York, 2nd edition, 2006. 3

[23] OpenPose. https://github.com/CMU-Perceptual-Computing-Lab/openpose. 2, 4, 5

[24] Paschalis Panteleris, Iason Oikonomidis, and Antonis Argyros. Using a single RGB frame for real time 3D hand pose estimation in the wild. In *WACV*, 2018. 1, 2

[25] Adam Paszke, Sam Gross, Soumith Chintala, Gregory Chanan, Edward Yang, Zachary DeVito, Zeming Lin, Alban Desmaison, Luca Antiga, and Adam Lerer. Automatic differentiation in pyTorch. In *NIPS-W*, 2017. 6

[26] Kathleen M. Robinette, Sherri Blackwell, Hein Daanen, Mark Boehmer, Scott Fleming, Tina Brill, David Hoeferlin, and Dennis Burnsides. Civilian American and European Surface Anthropometry Resource (CAESAR) final report. Technical Report AFRL-HE-WP-TR-2002-0169, US Air Force Research Laboratory, 2002. 4

[27] Matthias Teschner, Stefan Kimmerle, Bruno Heidelberger, Gabriel Zachmann, Laks Raghupathi, Arnulph Fuhrmann, Marie-Paule Cani, François Faure, Nadia Magnenat-Thalmann, Wolfgang Strasser, and Pascal Volino. Collision detection for deformable objects. In *Eurographics*, 2004. 1

[28] Dimitrios Tzionas, Luca Ballan, Abhilash Srikantha, Pablo Aponte, Marc Pollefeys, and Juergen Gall. Capturing hands in action using discriminative salient points and physics simulation. *IJCV*, 118(2):172–193, 2016. 1, 2



\end{document}